\documentclass[runningheads]{llncs}
\usepackage{graphicx}
\usepackage{tikz}
\usepackage{comment}
\usepackage{amsmath,amssymb} % define this before the line numbering.
\usepackage{color}

\newcommand{\etal}{\textit{et al}.}
\newcommand{\ie}{\textit{i}.\textit{e}.}
\newcommand{\eg}{\textit{e}.\textit{g}.}
\newcommand{\etc}{\textit{etc}.}

\usepackage{multirow}
\usepackage{siunitx}
\usepackage{makecell}
\usepackage{booktabs}
\usepackage{hyperref}
\hypersetup{hypertex=true,
            colorlinks=true,
            linkcolor=blue,
            anchorcolor=blue,
            citecolor=blue}
\usepackage{bbding}
\usepackage[width=122mm,left=12mm,paperwidth=146mm,height=193mm,top=12mm,paperheight=217mm]{geometry}

\begin{document}

\pagestyle{headings}
\mainmatter

\title{Towards Real-World Video Deblurring by Exploring Blur Formation Process} % Replace with your title

% CAMERA READY SUBMISSION
% \begin{comment}
\titlerunning{Real-world Video Deblurring}
% If the paper title is too long for the running head, you can set
% an abbreviated paper title here
%
\author{Mingdeng~Cao\inst{1} \and
Zhihang~Zhong\inst{2} \and
Yanbo~Fan\inst{3} \and
Jiahao~Wang\inst{1} \and
Yong~Zhang\inst{3} \and
Jue~Wang\inst{3} \and
Yujiu~Yang\inst{1}\href{mailto:yang.yujiu@sz.tsinghua.edu.cn}{\Envelope} \and
Yinqiang~Zheng\inst{2}\href{mailto:yqzheng@ai.u-tokyo.ac.jp}{\Envelope}}
\authorrunning{Cao et al.}
% First names are abbreviated in the running head.
% If there are more than two authors, 'et al.' is used.
%
\institute{
Tsinghua Shenzhen International Graduate School, Tsinghua University, China \and
The University of Tokyo, Japan \and
Tencent AI Lab, China
}
% \end{comment}
%******************
\maketitle

\begin{abstract}
This paper aims at exploring how to synthesize close-to-real blurs that existing video deblurring models trained on them can generalize well to real-world blurry videos. In recent years, deep learning-based approaches have achieved promising success on video deblurring task. However, the models trained on existing synthetic datasets still suffer from generalization problems over real-world blurry scenarios with undesired artifacts. The factors accounting for the failure remain unknown. Therefore, we revisit the classical blur synthesis pipeline and figure out the possible reasons, including shooting parameters, blur formation space, and image signal processor~(ISP). To analyze the effects of these potential factors, we first collect an ultra-high frame-rate (940 FPS) RAW video dataset as the data basis to synthesize various kinds of blurs. Then we propose a novel realistic blur synthesis pipeline termed as RAW-Blur by leveraging blur formation cues. Through numerous experiments, we demonstrate that synthesizing blurs in the RAW space and adopting the same ISP as the real-world testing data can effectively eliminate the negative effects of synthetic data. Furthermore, the shooting parameters of the synthesized blurry video, \eg, exposure time and frame-rate play significant roles in improving the performance of deblurring models. Impressively, the models trained on the blurry data synthesized by the proposed RAW-Blur pipeline can obtain more than 5dB PSNR gain against those trained on the existing synthetic blur datasets. We believe the novel realistic synthesis pipeline and the corresponding RAW video dataset can help the community to easily construct customized blur datasets to improve real-world video deblurring performance largely, instead of laboriously collecting real data pairs.

\end{abstract}

\section{Introduction}
\label{sec:introduction}
Video deblurring aims at restoring the latent sharp frame from the blurry input frames and has received considerable research attention. In which, deep neural networks~(DNN)-based deblurring approaches~\cite{su2017deep,nah2017deep,zhou2019spatio,wang2019edvr,li2021arvo,cao2022vdtr,zhong2022animation,wang2022efficient} are in the leading positions in recent years. In order to benchmark the video deblurring performance and facilitate the development of DNN-based models, large-scale and high-quality datasets are required. Therefore, developing an effective and convenient pipeline to create blurry-sharp pairs is indispensable.

% ----- results on BSD test dataset ---------
\begin{figure*}[!t]\footnotesize
    \centering
    \includegraphics[width=\linewidth]{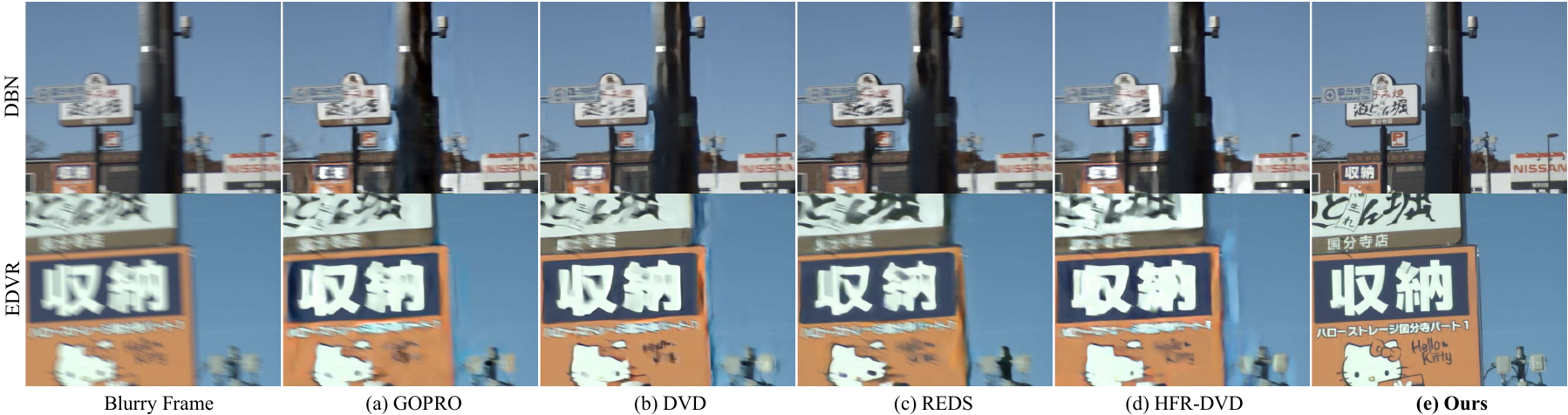}
    \vspace{-5mm}
    \caption{The real-world deblurring results of DBN~\cite{su2017deep} and EDVR~\cite{wang2019edvr}. The blurry videos come from the testing subset in the real-world video deblurring dataset BSD~\cite{zhong2020efficient}, and the models trained on the popular synthetic datasets, including GORRO~\cite{nah2017deep}, DVD~\cite{su2017deep}, REDS~\cite{nah2019ntire}, and HFR-DVD~\cite{li2021arvo}, cannot obtain satisfying results.}
    \label{fig:motivation}
    \vspace{-5mm}
\end{figure*}

To construct realistic blur datasets, most recently, researchers~\cite{rim2020real,zhong2020efficient} designed specific imaging systems that can record the blur-sharp pairs simultaneously. However, these systems require extremely high physical precision to align the blurry and sharp image acquisition cameras. Otherwise, the collected pairs may suffer from the misalignment problem easily. Meanwhile, these specific imaging systems are not flexible and can only capture sharp-blurry pairs with some fixed settings, which further restricts real-world deblurring performance. 

Compared to these real-world blur acquisition systems, synthesizing blurs is more convenient and flexible in constructing large-scale datasets for DNN-based model training. A classical approach is to convolve the sharp image with predefined motion blur kernels~\cite{kupyn2018deblurgan,shen2018deep,kaufman2020deblurring}. Yet, the synthesized blurs are unnatural since the bur kernels can only simulate limited uniform camera motions. A more realistic and popular way is to imitate the real-world blur formation process, and some researchers have proposed various large-scale synthetic deblurring datasets, such as GOPRO~\cite{nah2017deep}, DVD~\cite{su2017deep}, REDS~\cite{nah2019ntire}, and HFR-DVD~\cite{li2021arvo}. These datasets usually capture high-frame-rate sharp videos and then average several consecutive sharp frames to generate blurry frames blindly. 
Unfortunately, these blur synthesis pipelines are still quite different from the real-world blur formation process, including blur formation space, image signal processing, noise level, \etc We test some models trained on these popular synthetic datasets on the real-world blurry scenes in~\cite{zhong2020efficient} shown in Fig.~\ref{fig:motivation}. We can see that these models cannot effectively remove the blurs and may even destroy the image contents since the real-world blurs have a completely different blur formation process compared to existing synthetic blur pipelines. Thus developing a convenient pipeline is urgently needed to help the community synthesize realistic blurs to improve real-world video deblurring performance.

To move beyond aforementioned drawbacks of existing video deblurring datasets, in this work, 1) we aim at figuring out the intrinsic reasons why the models trained on existing video deblurring datasets cannot generalize well to the real-world blurry scenes, 2) and how to eliminate this gap on the data side. To this end, we firstly revisit existing blur synthesis pipelines by comparing them to the real-world blur formation process in detail. We find that the frame-rate of the sharp videos, blur formation space, noise, image signal processor, and blurry video parameters are the main factors attributing to the gap. Secondly, we collect a series of ultra high frame-rate RAW sharp videos so that we can synthesize various blurs to analyze and validate the factors effectively. Meanwhile, based on these RAW videos, we propose a novel pipeline termed RAW-Blur to synthesize realistic blurs for video deblurring. Thus existing models trained on these data can remove the blurs in the real world. Our RAW-Blur pipeline consists of three parts. Firstly, we should ensure the synthesized blurry video parameters~(\eg, exposure time and frame-rate) match the real blurry scene as closely as possible. Secondly, we average sharp frames in RAW space and add the reduced noise during the average process to simulate the real blur formation process. Last, the synthesized RAW blurs are transformed into RGB counterparts by a specific ISP close to the one adopted by the captured real blurry video. The whole pipeline is very close to the natural blur formation process, enabling handling blurs in the real world. Thus the models trained on the blur data synthesized by RAW-Blur can remove the real-world blurs shown in Fig.~\ref{fig:motivation}(e).

Our contributions can be summarized as follows~\footnote{This work was partially done when Mingdeng was an intern in Tencent AI Lab.}:
\begin{itemize}
    \item We revisit existing blur synthesis pipeline, and figure out the intrinsic reasons for the performance gap between synthetic and real blurs.
    \item We propose a novel realistic blur synthesis pipeline RAW-Blur, which can synthesize close-to-real blurs in RAW space. Meanwhile, we contribute an ultra-high frame-rate RAW video dataset to help the community construct their customized blur datasets.
    \item The quantitative and qualitative experimental results on the real-world blurry scenarios show the excellent real-world video deblurring performance of the proposed pipeline against existing blur synthesis methods.

\end{itemize}

% --------------------------------------------
\section{Related Works}
\label{sec:related}
\subsection{Deep Deblurring}
\label{sec:related_deblurring}
Recently, deep learning-based image and video deblurring approaches have achieved significant progress through more efficient spatial and temporal modeling. 

As for image deblurring, which tries to restore the clear image through effective spatial modeling, researchers usually design some specific architectures to enable a large receptive field. Nah~\etal~\cite{nah2017deep} proposed a multi-scale network to restore the clear images in a coarse-to-fine strategy. Based on this, Tao~\etal~\cite{tao2018scale} further shared the parameters of the restoration network among all scales, which reduces the parameters largely. Meanwhile,~\cite{zhang2019deep,suin2020spatially} utilized the hierarchical multi-patch network to deal with blurry images, achieving 40$\times$ faster runtime compared to previous multi-scale-based methods. Yuan~\etal~\cite{yuan2020efficient} also utilized deformable convolution~\cite{dai2017deformable} with optical flow guided training for spatially variant deconvolution.

Compared to image deblurring, video deblurring~(or multi-image deblurring) requires additional efficient temporal modeling to obtain satisfying results. Inspired by the encoder-decoder network, Su~\etal~\cite{su2017deep} designed a deep video deblurring network that directly outputs the central clear frame by concatenating consecutive frames as the model's input. To exploit the temporal information better, researchers adopt 3D convolution and recurrent neural network architecture for temporal modeling. However, the 3D-CNN-based model~\cite{zhang2018adversarial} requires huge computational costs, and simple RNN-based models~\cite{hyun2017online,nah2019recurrent,zhong2020efficient} still are inefficient for temporal modeling. Based on these models, some temporal alignment and aggregation methods are proposed to model the temporal variations. Wang~\etal~\cite{wang2019edvr} utilized deformable convolution, and Zhou~\etal~\cite{zhou2019spatio} adopted dynamic convolution to align neighboring frames, achieving better performance with moderate parameters. Meanwhile, optical flow is further adopted to align multiple frames to aggregate complementary information in~\cite{pan2020cascaded,son2021recurrent,li2021arvo}.

\subsection{Motion Blur Datasets}
\label{sec:related_datasets}
In the early days, the research community applied blur kernels to synthesized blurry images, which usually convolve the sharp images with predefined uniform blur kernels and add Gaussian noise. Since assuming a uniform blur over the image is not realistic even for planar scenes, Levin~\etal~\cite{levin2009understanding} locked the tripod's Z-axis rotation handle but release the X and Y handles to create eight spatially invariant blur kernels. The maximum collected kernel size is up to $41\times 41$. Then, Lai~\etal~\cite{lai2016comparative} utilized the algorithm in~\cite{schmidt2013discriminative} to synthesize 4 larger blur kernels, ranging from $51\times 51$ to $101\times 101$ by randomly sampling 6D camera trajectories. More recently, Shen~\etal~\cite{shen2018deep} synthesized 20,000 motion blur kernels with sizes ranging from $13\times 13$ to $27\times 27$ based on~\cite{boracchi2012modeling}'s random 3D camera trajectories to generate a blurred face dataset. However, images with uniform blur are different from real captured cases which usually have spatially varying blurs.

The blurring process can be simulated by the integration of sharp images during the shutter time. To generate non-uniform blurs and reduce the gap between synthetic and real-world blurry images, researchers resort to average consecutive frames from high frame-rate videos to synthesize blur-sharp image/video pairs. Su~\etal~simulated a motion blur dataset, dubbed as DVD~\cite{su2017deep}, at \SI{30}{FPS} by sub-sampling every eighth frame of \SI{240}{FPS} sharp videos from various devices, including iPhone 6s, GoPro Hero 4 Black, and Canon 7D. The blurry images are synthesized by averaging 7 nearby frames. To alleviate the ghosting artifacts (\eg, unnatural spikes or steps in the blur trajectory) caused by adjacent exposures,~\ie, duty cycle, Su~\etal~computed optical flow between adjacent frames and generate extra 10 evenly spaced inter-frame images before averaging. Meanwhile, Nah~\etal~proposed a motion blur dataset GOPRO~\cite{nah2017deep}, using \SI{240}{FPS} videos from GOPRO4 Hero Black camera. Instead of averaging consecutive frames directly in RGB space, GOPRO~\cite{nah2017deep} applies an inverse gamma function to the images and then averages them to avoid the effects caused by nonlinear CRFs. Then, Nah~\etal~created a large dataset REDS~\cite{nah2019ntire} by incorporating the experiences from DVD~\cite{su2017deep} and GOPRO~\cite{nah2017deep}. The difference is that REDS~\cite{nah2019ntire} directly uses the off-the-shelf video-frame-interpolation model~\cite{niklaus2017video} to interpolate the videos from \SI{120}{FPS} to \SI{1920}{FPS} before averaging. Nevertheless, the current deep learning-based models trained on previous synthetic datasets still have poor generalization on real-world samples. 

Recently, some researchers have started to shoot the real-world blur-sharp image pairs directly by designing a precise beam-splitter acquisition system. Specifically, two cameras are physically aligned to receive the light split in half by the beam splitter. With one camera using a long exposure and the other using a short exposure, the acquisition system can take both sharp and blurry image pairs of the same scenarios simultaneously. Based on the above devices, Rim~\etal~and Zhong~\etal~proposed the first real-world motion blur image dataset RealBlur~\cite{rim2020real} and video dataset BSD~\cite{zhong2020efficient}, respectively. Zhong~\etal~further proposed a dataset BS-RSCD~\cite{zhong2021towards} for the joint rolling shutter deblurring and correction task. However, real-world datasets have very high precision requirements for the beam-splitter acquisition system, otherwise it is very easy to cause image misalignment. For shooting videos with different devices or different exposure parameters, it is cumbersome and risky to use such equipment.

Although these datasets promote the development of the deblurring significantly, the models trained on these datasets still cannot generalize well to real-world blurry scenarios. 

\section{Blur Formation Process Revisit}
\label{sec:pipelines}
\subsection{Real-World Blur Formation}
\label{sec:real_blur_accumulation}
\begin{figure}[!t]
    \centering
    \includegraphics[width=0.99\linewidth]{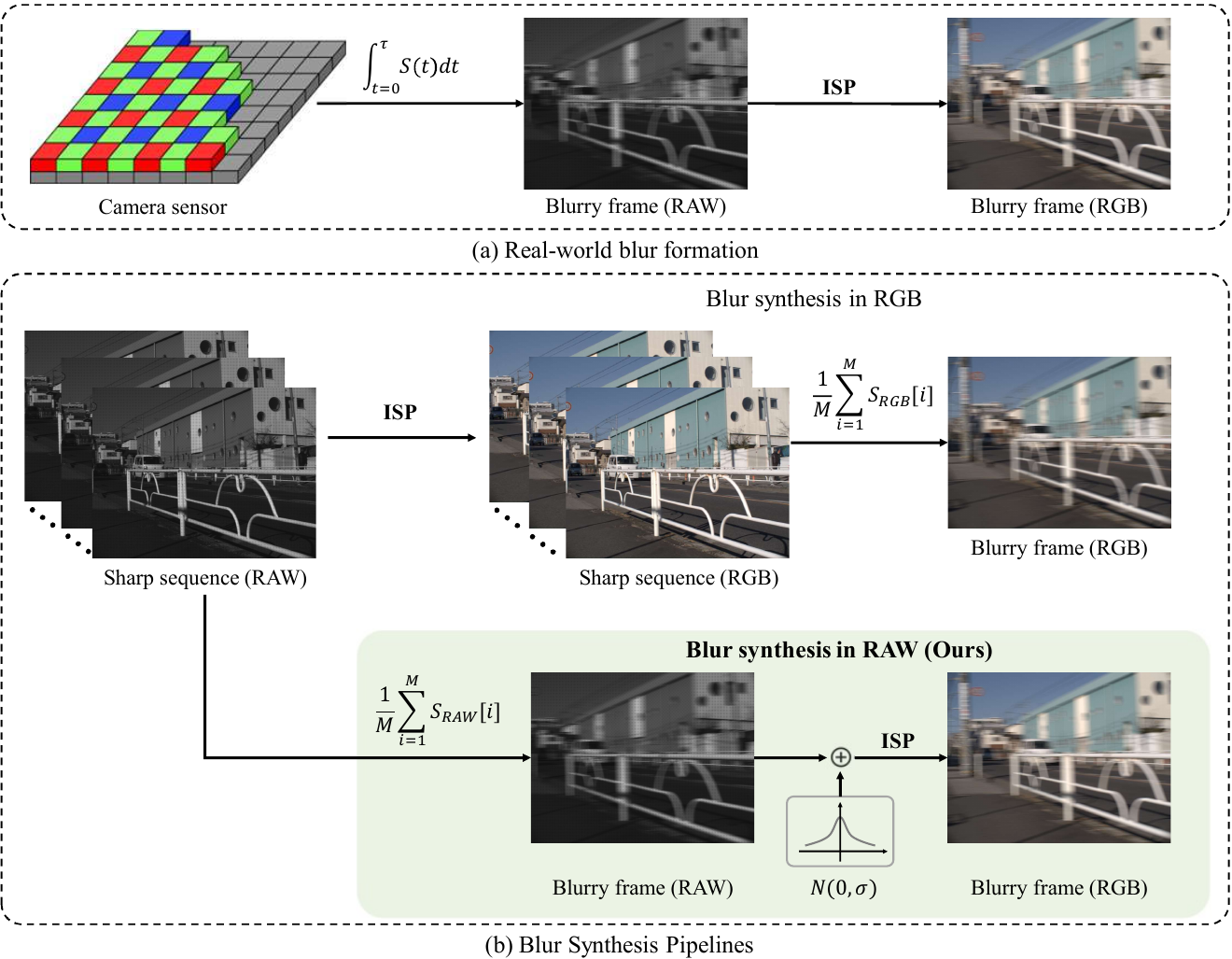}
    \vspace{-5mm}
    \caption{Real-world and synthetic blur formation processes. Our pipeline directly synthesize the blurs in RAW space and further add the noise to simulate the real blurs.}
    \label{fig:pipelines}
\vspace{-5mm}
\end{figure}

We simplify the typical digital camera imaging process into two key steps: image signal acquisition~(RAW data) and RGB rendering with ISP. As shown in Fig.~\ref{fig:pipelines}(a), during the signal acquisition, the input lights are captured and transformed into digital signals by the camera sensor with a series of camera settings, \eg, exposure time and frame-rate. During this process, the blurs occur when relative movements arise during the exposure time. This real blur accumulation process can be formulated as:
\begin{equation}
    \label{eq:real_blur_formation}
    B_{real} = \int_{0}^{\tau}S(t)dt,
\end{equation}
where $B_{real}$ and $S(t)$ are the blurry RAW frame and the signal captured by the camera sensor at time $t$, respectively. $\tau$ is the exposure time. After that, these signals~(RAW image) are rendered into RGB format with camera ISP, including white balance, demosaicing, gamma correction, color correction, \etc. 

\subsection{Existing Blur Synthesis Pipeline}
\label{sec:exist_datasets}
Existing synthetic blur datasets, \eg, GOPRO~\cite{nah2017deep}, DVD~\cite{su2017deep}, REDS~\cite{nah2019ntire}, and HFR-DVD~\cite{li2021arvo}, are constructed in a blind way that discretizes the continuous blur accumulation process (described in~Eq.~\ref{eq:real_blur_formation}) by averaging consecutive frames of sharp videos. In practice, researchers first record the high frame-rate observed~(in RGB format) videos directly, then DVD and HFR-DVD average these observed frames to synthesize blurs:
\begin{equation}
    \label{eq:discrete_blur}
    B_{syn} = \frac{1}{M} \sum_{i=1}^{M}\hat{S}[i],
\end{equation}
where $B_{syn}$ is the synthesized blurry frame, $ M$ and $\hat{S}[i]$ are the number of sampled frames, and the $i-$th observed sharp frame during the exposure time. These synthesized blurs cannot generalize well in real-world blurs since the RGB space is non-linear, and averaging in this space may generate different blur patterns against real-world blurs. GOPRO and REDS further adopt the inverse CRF to transform the observed sharp frames into linear space and then synthesize blurs~\cite{nah2017deep}:
\begin{equation}
    \label{eq:discrete_blur_crf}
    B_{syn} = g(\frac{1}{M} \sum_{i=1}^{M}g^{-1}(\hat{S}[i])),
\end{equation}
where $g$ is the non-linear CRF. In this way, the blur synthesis process is closer to the natural blur accumulation process. Yet, they still suffer from domain gaps since the CRF is unknown and hard to estimate accurately. Meanwhile, DVD and REDS utilize frame interpolation to increase video continuity; thus, more frames interpolated during inter-frame time generate more continuous blur. To be convenient, we summarize the detailed configurations and the critical operations in blur synthesis of these datasets in Tab.~\ref{tab:existing_datasets}. 

\setlength{\tabcolsep}{4pt}

\begin{table}[htbp]
\vspace{-5mm}
    \centering
    \caption{The detailed configurations of existing synthetic video deblurring datasets}
    \label{tab:existing_datasets}
    \resizebox{0.9\linewidth}{!}{
    \begin{tabular}{lcccc}
    \toprule
    \noalign{\smallskip}
    Datasets     & GOPRO~\cite{nah2017deep} & DVD~\cite{su2017deep} & REDS~\cite{nah2019ntire} & HFR-DVD~\cite{li2021arvo} \\
    \noalign{\smallskip}
    \hline
    \noalign{\smallskip}
    Sharp FPS & 240 & 240& 120 & 1000 \\
    Blurry FPS    & $\approx$25 & $\approx$30 & 24  &  25  \\
    \hline
    Frame Interpolation &  & \checkmark & \checkmark &  \\
    Inverse CRF Calibration  & \checkmark &  & \checkmark &  \\
    \bottomrule
    \end{tabular}
    }
\vspace{-10mm}
\end{table}

\subsection{Synthetic Blur Analysis}
\label{sec:factor_analysis}
In the following, we analyze the intrinsic reasons that the models trained on these synthetic datasets cannot generalize well on real-world blurry scenes by comparing synthetic blurs to real ones.

\begin{figure*}[t]
    \centering
    \includegraphics[width=0.95\linewidth]{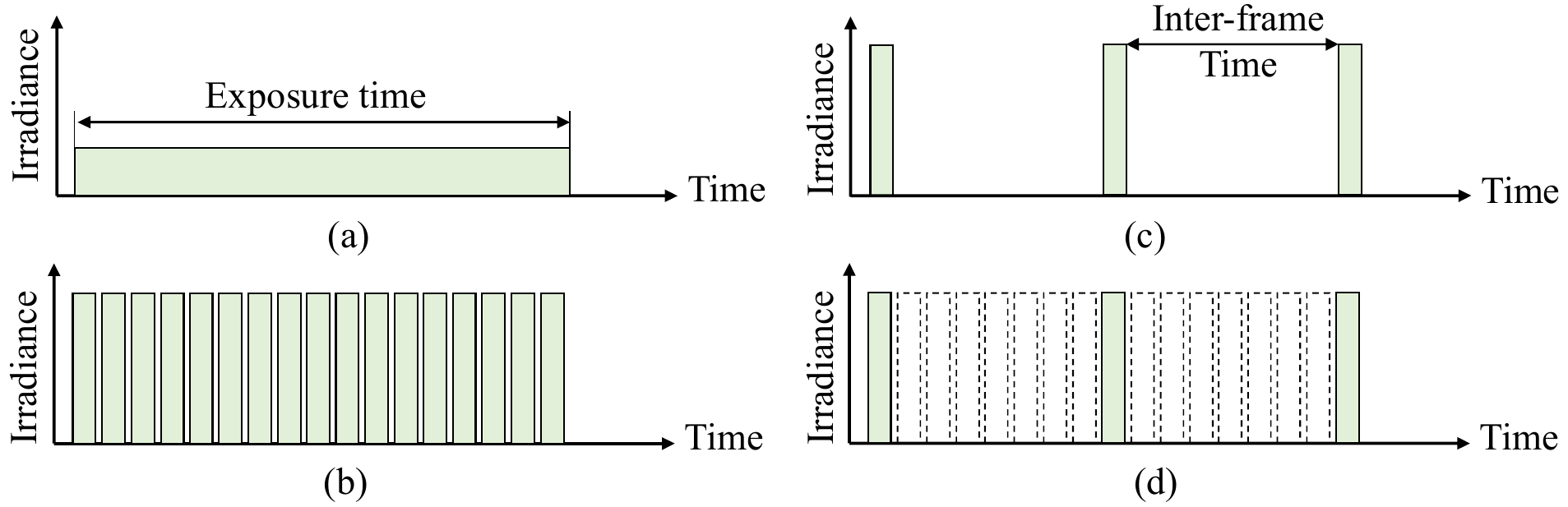}
    \vspace{-5mm}
    \caption{Exposure strategies of real-world blurry and sharp videos acquisition process. \textbf{(a)} Long exposure time during real-world blur formation. \textbf{(b)} Very short exposure time during higher frame-rate sharp video capture. \textbf{(c)} Lower frame-rate sharp frames exposure strategy. \textbf{(d)} Lower frame-rate sharp frames with frame interpolation }
    \label{fig:framerate}
\vspace{-5mm}
\end{figure*}

\vspace{2pt}\noindent\textbf{High Frame-rate Sharp Video. }
\label{para:continuity}
To simulate the real blur formation process (shown in Fig.~\ref{fig:framerate}(a)), short exposure time and high frame-rate are required to acquire sharp videos. However, existing synthetic datasets usually capture sharp videos at 120 FPS~(\eg, REDS) or 240 FPS~(\eg, GOPRO, DVD), which own large inter-frame time~(shown in Fig.~\ref{fig:framerate}(c)), resulting in synthesizing unreal blurs that are discontinuous and suffer from undesired spikes. Accordingly, DVD and REDS try to alleviate this problem by further increasing frames during the inter-frame time with frame interpolation techniques~(shown in Fig~\ref{fig:framerate}(d)). Nevertheless, the interpolated videos suffer from artifacts due to the occlusions~\cite{bao2019memc}. Thus higher frame-rate sharp videos with short inter-frame time are needed for more continuous and close-to-real blurs synthesis.

\vspace{2pt}\noindent\textbf{Blur Formation Space.}
\label{para:space}
We should reclaim that the real-world blurs are formatted as RAW signals rather than the observed RGB images. Existing synthetic deblurring datasets synthesize blurs in the RGB space directly, \eg, DVD~\cite{su2017deep} and HFR-DVD~\cite{li2021arvo}, deteriorating the real-world deblurring performance significantly and introducing undesired artifacts. Though GOPRO~\cite{nah2017deep} and REDS~\cite{nah2019ntire} proposed to synthesize blurs with the observed images calibrated by the inverse CRF, the CRF is hard to estimate accurately. Thus realistic blurs should be synthesized from RAW frames instead of RGB frames.

\vspace{2pt}\noindent\textbf{Imaging Noise. }
Some noises are inevitable during the imaging process~\cite{liu2006noise,foi2008practical}, especially when blurs occur, where the exposure time is relatively long in low-light conditions. However, when we adopt the pipeline described in~Sec.~\ref{sec:exist_datasets} to synthesize blurs, the noise tends to be eliminated for the following reasons. Firstly, the captured sharp videos are of different levels of imaging noise because of the short exposure time compared to blurry videos. Secondly, averaging multiple frames can reduce the noise largely~(please refer to the proof in the provided supplementary materials). Yet, existing blur synthesis pipelines ignore the noise, widening the gap between these synthetic and real-world blurs. Thus, it is reasonable to consider additional noise to generate more realistic blurs.

\vspace{2pt}\noindent\textbf{Image Signal Process~(ISP). } 
Camera ISPs transform the captured RAW data into RGB images. In this process, different camera manufacturers develop their own processing pipelines based on the traditional ISP algorithms. Therefore, the blurs synthesized from the sharp frames captured by one camera may not generalize well to the blurry scenes recorded by the cameras adopting another ISPs. Thus we should further consider the impacts of different ISPs in removing real-world blurs. 

\vspace{2pt}\noindent\textbf{Video Parameter Setting. }
As for video deblurring, modeling on the temporal variation is significant. The exposure time and frame-rate of the synthesized blurry videos are two critical factors. Specifically, the exposure time highly relates to the blurring degrees, and the frame-rate is essential for temporal modeling. Thus we can consider that the difference in terms of these two factors between synthetic and real-world blurry videos is also one of the reasons why the real-world video deblurring cannot be resolved well by existing  synthetic blur datasets~\cite{nah2017deep,su2017deep,nah2019ntire,li2021arvo}.

\section{RAW-Blur Dataset Construction}
\label{sec:dataset}
In this section, we elaborate on synthesizing a blur dataset for real-world video deblurring considering the factors described in Sec.~\ref{sec:pipelines}. We propose a novel synthesis pipeline called RAW-Blur, which can generalize well in real blurry videos. In brief, we first collected an ultra-high frame-rate RAW video dataset named UHFRaw. Then, we synthesize realistic blurs based on the proposed UHFRaw.

\subsection{Ultra-High Frame-rate RAW Video Dataset}
We first collect an high frame-rate raw video dataset termed as UHFRaw. The videos are captured at a ultra-high frame-rate with the RAW format. Specifically, we collect 45 videos of resolution $812\times 620$ at 940 FPS totally, by a BITRAN CS-700C camera. Each video is about 3 seconds long and contains about 2900 frames. Due to the high-frequency exposure strategy, the exposure time is approximately 1.01ms and inter-frame time is nearly zero. Therefore, the captured raw frames are sharp and continuous enough to synthesize realistic blurs without frame interpolation and inverse CRF estimation. The scenes in the captured videos are mainly street scenes, where various vehicles, pedestrians are included.

We also note that a RAW dataset named Deblur-RAW is also proposed for image deblurring in~\cite{liang2020raw}. Compared to our UHFRaw, they only record the RAW videos at a low frame-rate~(30 FPS) and average 3-5 frames to synthesize blurs. Therefore, the synthesized blurs are unreal and unnatural, further introducing undesired artifacts~(please refer to our supplementary material).

\subsection{RAW-Blur Synthesis Pipeline}
\label{sec:raw_blur}
Our blur synthesis pipeline for real-world video deblurring is shown in Fig.~\ref{fig:pipelines}, which contains three main parts. First, we should set the exposure time and the frame-rate of the synthesized video. Then, we synthesize the RAW blurry images by averaging consecutive RAW sharp frames, and further add the reduced noise to the blurry RAW frame, making it closer to the real blurry scenes. Last, the noisy blurry RAW frames are rendered into RGB format by the Camera ISP.

\vspace{2pt}\noindent\textbf{Blurry Video Parameter Setting.}
As our aim is to synthesize a realistic dataset for video deblurring, we should consider the frame-rate of the video and exposure time. We denote the time of each blurry frame~(inverse of frame-rate) and its exposure time as $T$-ms and $\tau$-ms. Thus the dead-time between consecutive frames is ($T - \tau$)-ms, and each frame owns the duty cycle $\frac{\tau}{T}$. Since our UHFRaw the exposure time of each frame in UHFRaw is nearly $1$-ms, thus $T$ and $\tau$ are approximately equal to the number of frames. We can easily synthesize blurry videos with various video parameters by changing these two hyper-parameters. Our experiments in Sec.~\ref{sec:experiments} demonstrate the impact of the synthesized video parameters for real-world video deblurring.

\vspace{2pt}\noindent\textbf{RAW-Blur Synthesis.}
Instead of averaging consecutive frames in the RGB space adopted by existing blur synthesis methods, we average the sharp high frame-rate RAW frames directly to synthesize more realistic blurs blindly~(the detailed comparison is shown Fig.~\ref{fig:pipelines}(b)). Meanwhile, as we analyzed in Sec.~\ref{sec:factor_analysis}, the noise in the sharp frames is corrupted significantly when we adopt averaging consecutive frames strategy to simulate the blurring process. To synthesize close-to-real blurs in the wild, we add the reduced Gaussian noise into the synthesized RAW blurry frame. Thus the whole RAW blur synthesis pipeline can be formulated as:
\begin{equation}
    B_{raw} = \frac{1}{\tau}\sum_{i=0}^{\tau} S_{raw}[i] + N(0, \sigma),
\end{equation}
where $B_{raw}$ is the synthesized RAW blurry frame, and $S_{raw}[i]$ is the sharp RAW frame at index $i$. $N(0, \sigma)$ is the noise distribution with expectation $0$ and variance $\sigma$. In our pipeline, $\sigma$ consists of the signal-independent and dependent parts, \ie, the classical Poisson-Gaussian noise model.

\vspace{2pt}\noindent\textbf{Rendered with ISP.}
After synthesizing the RAW blurry frames with noise, we should translate them into RGB frames with the camera ISP for model training. Since each manufacturer designs its own specific and nonpublic ISP for processing the RAW images, the blurry videos rendered by different ISPs own different image properties, like brightness, color constancy. As a result, the models trained with the blurs rendered by one ISP cannot generalize on the blurs rendered by another. Therefore, we propose to apply the same ISP of the captured real-world blurry videos to transform the synthesized raw blurry frames into RGB blurry frames as far as possible.

Compared to the existing blur synthesis pipeline, our RAW-Blur synthesis pipeline has the following advantages that can generate close-to-real blurs: 1) RAW-Blur considers the effects of blurry video parameters for real-world video deblurring. 2) The ultra-high frame-rate sharp frames ensure the temporal continuity to synthesize natural and real blurs, avoiding the artifacts introduced by the interpolation operation. 3) Synthesizing blurs in RAW space and adding reduced noise is closest to the real blur formation process, while existing blur synthesis methods suffer from the non-linear property in RGB space. 4) The effect of camera ISP is further considered to improve existing models' real-world video deblurring performance.

% \vspace{-3mm}
\section{Experiments}
% \vspace{-3mm}
\label{sec:experiments}
\subsection{Experimental Setting}
\noindent\textbf{Datasets.}
We compare the proposed realistic blur synthesis pipeline to the popular synthetic datasets, including GOPRO, DVD, REDS, and HFR-DVD. Meanwhile, to evaluate the real-world video deblurring performance quantitatively, we conduct validation on the recently proposed real-world video deblurring dataset BSD~\cite{zhong2020efficient}. There are three subsets with different exposure times for sharp and blurry image pairs in BSD, \ie, 1ms-8ms, 2ms-16ms, and 3ms-24ms, respectively. In the following, $T$ and $\tau$ are set as 33 and 11 in our synthesized datasets without special instructions.

\vspace{2pt}\noindent\textbf{Implementation Details.} 
We adopt a classical DNN-based video deblurring model DBN~\cite{su2017deep} and a state-of-the-art model EDVR~\cite{wang2019edvr} without any modifications to validate the effectiveness of the proposed blur synthesis pipeline on deblurring real-world blurry videos. During training, we randomly crop the input frames into $256 \times 256$ and randomly flip them both horizontally and vertically for the data augmentation. The initial learning rates are $4\times 10^{-4}$ and $10^{-4}$ for DBN and EDVR, respectively. ADAM~\cite{diederik2014adam} optimizer is employed to optimize the model's parameters. We train the models with 1000 epochs totally, and the learning rate decays 10 times at 700-, 850-, and 920-th epoch. For evaluation, both PSNR and SSIM~\cite{wang2004image} are adopted to evaluate the deblurring performance quantitatively. All codes to reproduce the results will be made public~\footnote{The project is available at~\url{https://github.com/ljzycmd/RAWBlur}.}.

\subsection{Comparison to Existing Synthetic Datasets}
\begin{figure}[tbp]\footnotesize
    \centering
    \includegraphics[width=0.95\linewidth]{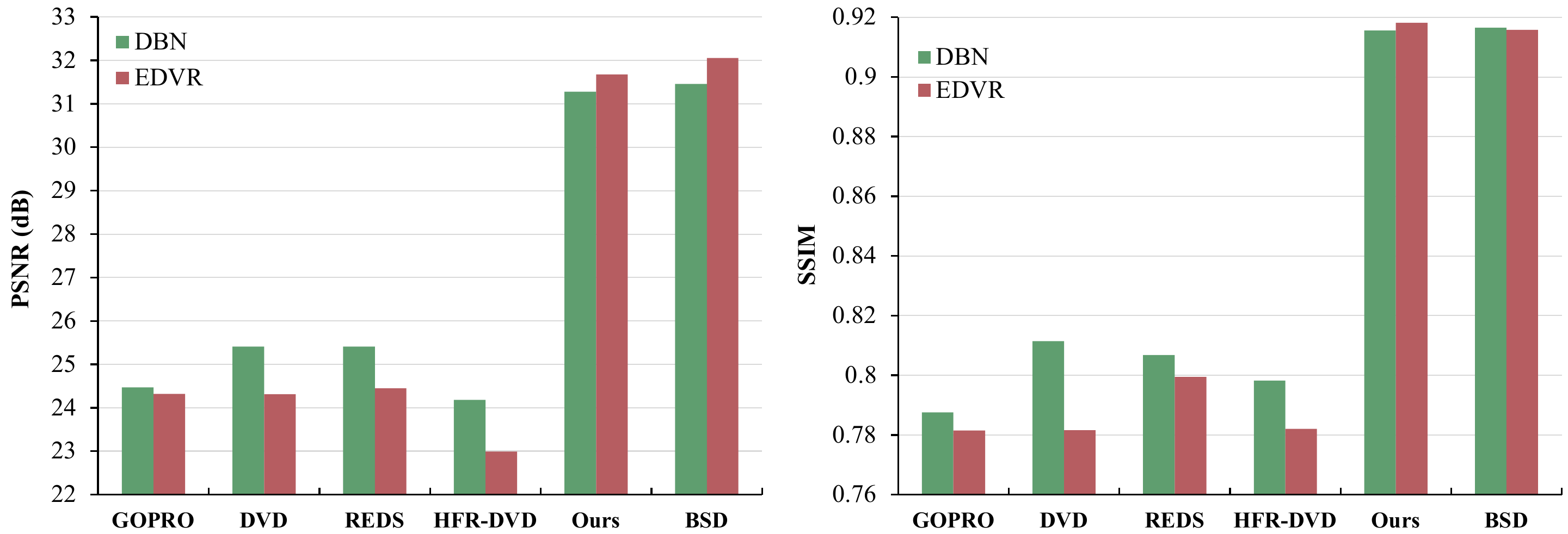}
    \vspace{-3mm}
    \caption{The performance~(\ie, PSNR and SSIM) comparison of the models trained on different synthetic datasets and real-world datasets in the testing subset of BSD. We can see that both DBN and EDVR obtain much higher PSNR when trained on our synthetic dataset and nearly the same PSNR as the models trained with BSD. This demonstrates our synthesis pipeline is closer to the real-world blur formation process. We further provide the PSNR and SSIM metrics in our supplementary materials.}
    \label{fig:performance_comparison}
\vspace{-5mm}
\end{figure}

The quantitative comparison between the dataset synthesized by our RAW-Blur pipeline and existing synthetic datasets is shown in Fig.~\ref{fig:performance_comparison}. We see that when trained on existing synthetic datasets, both DBN and EDVR obtain very low evaluation metrics since the blurs in these synthetic datasets deviate from real-world blurry scenes. On the contrary, these two models trained on our synthesized dataset achieve higher PSNR and SSIM with a large margin against other synthetic datasets. Besides, the metrics are much closer to those of the models trained on the training subset of BSD~\footnote{Note that the dataset synthesized by RAW-Blur only contains 45 videos with 3800 frames in total, which is 35\% less than the training data in BSD~\cite{zhong2020efficient}}. This reveals that synthesizing blurs in RAW space with noise is closer to the real blur formation process. 

\begin{figure*}[!ht]\footnotesize
    \centering
    \includegraphics[width=\linewidth]{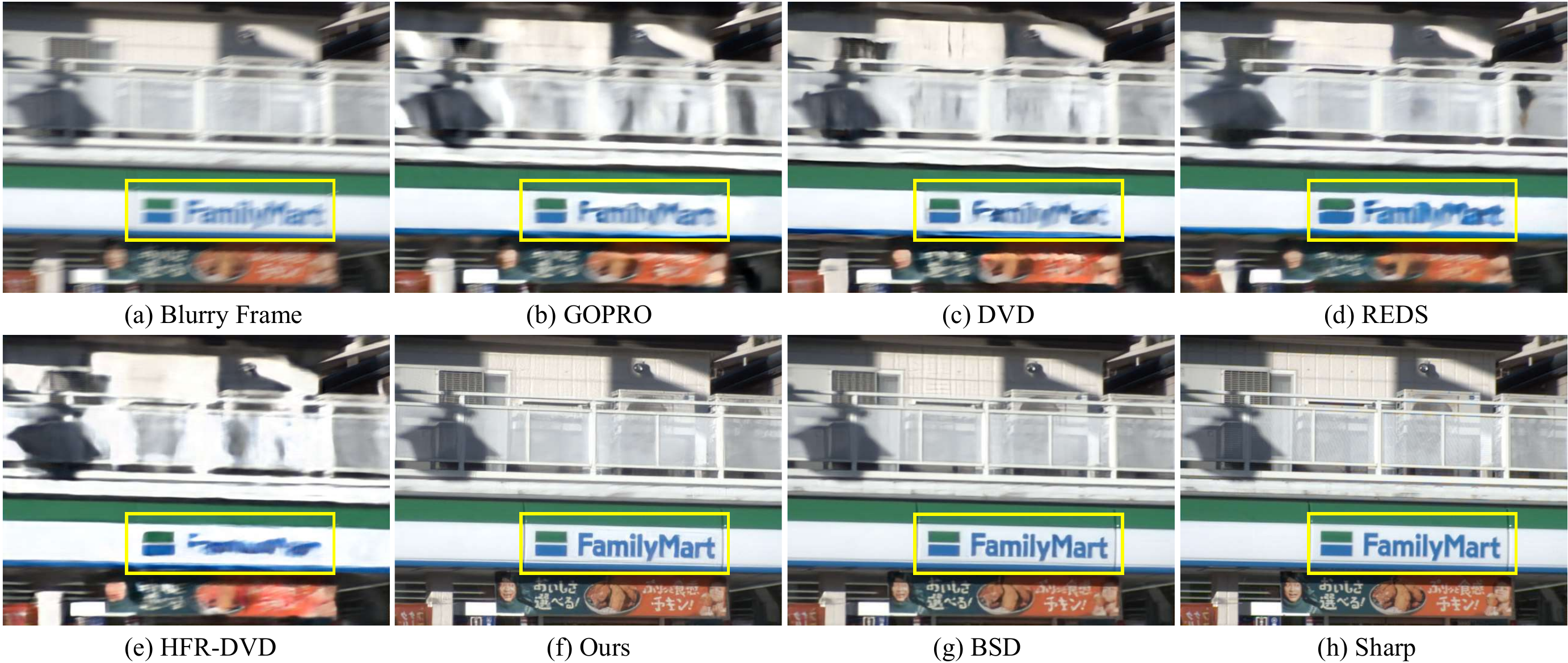}
    % \vspace{-5mm}
    \caption{Visual comparison of EDVR~\cite{wang2019edvr} on the real-world BSD testing subset. The model trained on the blur dataset synthesized by RAW-Blur shows a strong competitive edge against existing synthetic datasets. More results can be found in the supplementary materials.}
    \label{fig:visual_comparison}
\vspace{-5mm}
\end{figure*}

Meanwhile, we further visualize the deblurred results of EDVR trained on different datasets in Fig.~\ref{fig:visual_comparison}. We see that EDVR fails in dealing with real-world blurry scenes when trained on existing synthetic datasets, and our RAW blur pipeline can restore friendly visual results. These quantitative and qualitative results demonstrate the effectiveness of our blur synthesis pipeline, which can be used to remove real-world blurs.

% ----- Visual Comparison ---------

\vspace{-3mm}
\subsection{Analysis of RAW-Blur Synthesis Pipeline}
In the following, we analyze the effects of the key components mentioned in Sec.~\ref{sec:factor_analysis} for real-world video deblurring by conducting numerous comparative experiments corresponding to each factor.

\begin{figure*}[!t]\footnotesize
    \centering
    \includegraphics[width=\linewidth]{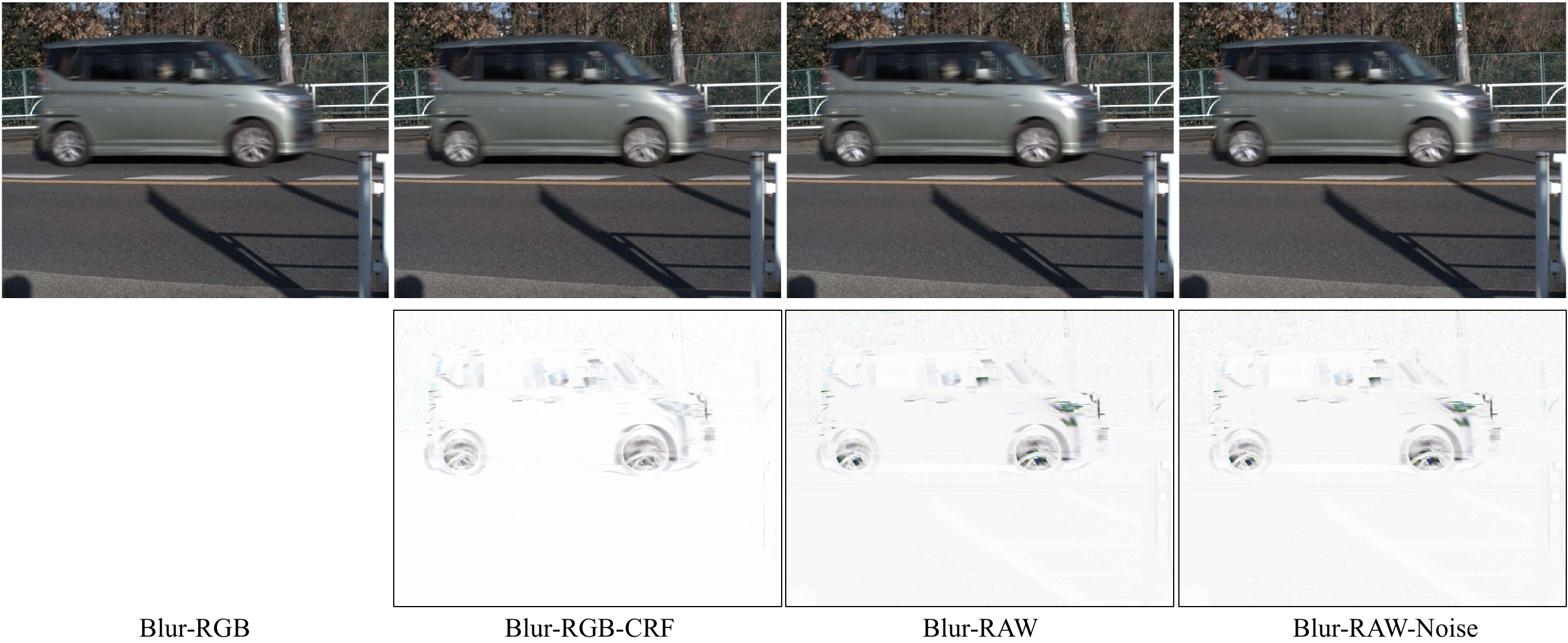}
    \caption{The visualization of the synthesized blurs in different space and the difference between the blurs synthesized in RGB space and RGB-CRF, RAW, RAW-Noise space. We see that these data synthesized in different spaces differ significantly mainly in the blurry areas, \eg, the blurry car.}
    \label{fig:difference}
\vspace{-3mm}
\end{figure*}

\vspace{2pt}\noindent\textbf{Blur Synthesis Space.}
We first conduct experiments of the blur synthesis space for real-world video deblurring, including RGB space adopted in DVD~\cite{su2017deep} and HFR-DVD, RGB-CRF space adopted in GOPRO~\cite{nah2017deep} and REDS~\cite{nah2019ntire}, and the proposed RAW and RAW-Noise~(RAW with noise) space. Thus we synthesize four sub-datasets by averaging consecutive sharp frames in these four spaces, and the difference of each strategy is visualized in Fig.~\ref{fig:difference}. The newly synthesized datasets are with the same setting except for the synthesis space, and the evaluation results of different strategies are shown in Tab.~\ref{tab:ab_factors}. We see that directly averaging in RGB space gets the lowest PSNR and SSIM, and the model's performance improves largely when synthesizing the blurs in RGB-CRF, for that the inverse CRF can transform the non-linear RGB images into nearly linear RGB images. And blur synthesis in the RAW space surpasses it since the RAW space is purely linear. When further adding noise into the synthesized RAW blurry images to imitate realistic blurs, we further improve EDVR's performance in terms of 0.3dB higher PSNR. 

These results demonstrate the importance of the synthesis space, and simply averaging frames cannot reflect the real-world blur formation process. Therefore, we recommend synthesizing realistic blurs in RAW space and adding noise. 

% ----- difference of different synthesis space ---------

\begin{table}[!h]
    \centering
    \caption{Effects of synthesis space, video frame interpolation and ISP. The performance~(PSNR/SSIM) are evaluated in the testing subset of BSD~\cite{zhong2020efficient}. Note that the frames in BSD are rendered by RawPy~\cite{maik2014rawpy}}
    \label{tab:ab_factors}
    \begin{tabular}{l@{\extracolsep{0.5cm}}l@{\extracolsep{0.5cm}}ccc}
    \toprule
    Methods & Synthesis Space & Interpolation & ISP & BSD (2-16 ms) \\
    \hline
    \multirow{8}{*}{DBN}
        & RGB       &         & RawPy & 29.10/0.8925  \\
        & RGB     &\checkmark & RawPy & 28.03/0.8695 \\
        & RGB-CRF   &         & RawPy & 30.67/0.9003 \\
        & RGB-CRF &\checkmark & RawPy & 28.19/0.8521 \\
        & RAW       &         & RawPy & 30.86/0.9025 \\
        & RAW     &           & DarkTable & 29.86/0.8887     \\
        & RAW-Noise &   &RawPy & {31.28/0.9156} \\
        \hline
    \multirow{8}{*}{EDVR}
        & RGB       &         & RawPy & 29.77/0.8893 \\
        & RGB     &\checkmark & RawPy & 27.91/0.8739 \\
        & RGB-CRF   &         & RawPy & 31.06/0.9059 \\
        & RGB-CRF &\checkmark & RawPy & 29.68/0.8812 \\
        & RAW       &         & RawPy & 31.39/0.9072 \\
        & RAW     &           & DarkTable & 29.68/0.8891 \\
        & RAW-Noise &   & RawPy & {31.68/0.9192} \\
    \bottomrule
    \end{tabular}
\vspace{-5mm}
\end{table}
\setlength{\tabcolsep}{4pt}

\vspace{2pt}\noindent\textbf{Ultra High Frame-rate Sharp Videos.}
We then validate the effectiveness of the high frame-rate sharp videos. To do so, we further synthesize the blurs with frame interpolation in RGB space since there are nearly no RAW video interpolation techniques. Specifically, we first temporally downsample the frame-rate 8 times to generate sharp videos at about 120 FPS. Then we adopt a state-of-the-art video interpolation model RIFE~\cite{huang2020rife} to interpolate the low frame-rate videos into 940 FPS. The quantitative results on the BSD are shown in Tab.~\ref{tab:ab_factors}. We see that the model trained on the interpolated data achieved 1dB lower PSNR than the one without interpolation. These results ravels that the interpolation may introduce some artifacts, and the synthesized blurs still own the domain gap compared to the real-world accumulated blurs. Thus, we highly recommend capturing native high FPS videos to synthesize realistic blurry frames. 

\vspace{2pt}\noindent\textbf{Different ISPs.}
We also adopt a free RAW image processing software named DarkTable~\cite{darktable2022} to render the synthesized RAW blurry frames to analyze the effect of different ISPs, shown in Tab.~\ref{tab:ab_factors}. Both DBN and EDVR suffer from 1.0dB and 1.7dB PSNR drop compared to the models trained over the frames rendered by RawPy~\cite{maik2014rawpy}. This reveals that the blurs rendered by different ISPs differ greatly, resulting in the models trained with the frames processed by one specific ISP cannot generalize well on the frames processed by another. Thus, we highly recommend adopting the same ISP of the real-world blurry scenes to render the synthesized RAW blurry frames for model training. 

% ----- different temporal strategies ------
\setlength{\tabcolsep}{4pt}
\begin{table}[t]
    \centering
    \caption{The quantitative evaluation results of different video parameters of the synthesized blurry video. Note that the FPS and exposure time of the synthesized blurry video are controlled by setting different $T$ and $\tau$. }
    \label{tab:shooting_param}
    \begin{tabular}{llcccc}
    \toprule
    \noalign{\smallskip}
    Methods & FPS  & Exposure Time & BSD (1-8 ms) & BSD (2-16 ms) & BSD (3-24 ms) \\
    \noalign{\smallskip}
    \hline
    \noalign{\smallskip}
    \multirow{5}{*}{DBN}
        & $\approx$15  &  $\approx$8ms  & 31.65/0.9151 & 30.72/0.8983 & 30.05/0.8971  \\
        & $\approx$15  &  $\approx$16ms & 31.33/0.9105 & 30.77/0.9018 & 30.60/0.9113 \\
        & $\approx$15  &  $\approx$24ms & 30.70/0.9007 & 29.69/0.8816 & 30.07/0.8999 \\
        & $\approx$28  &  $\approx$11ms & 31.22/0.9094 & 30.87/0.9019 & 30.58/0.9083 \\
        & $\approx$43  &  $\approx$11ms & 29.37/0.8912 & 29.12/0.8839 & 30.26/0.9048 \\
    \noalign{\smallskip}
    \hline
    \noalign{\smallskip}
    \multirow{5}{*}{EDVR}
        & $\approx$15  &  $\approx$8ms  & 32.02/0.9213 & 31.62/0.9129 & 31.05/0.9153  \\
        & $\approx$15  &  $\approx$16ms & 32.04/0.9200 & 31.76/0.9129 & 31.81/0.9274 \\
        & $\approx$15  &  $\approx$24ms & 30.67/0.9072 & 31.04/0.9047 & 31.60/0.9235 \\
        & $\approx$28  &  $\approx$11ms & 29.85/0.9043 & 31.62/0.9109 & 28.91/0.9053 \\
        & $\approx$43  &  $\approx$11ms & 27.49/0.8555 & 27.22/0.8443 & 28.62/0.8763 \\
    \noalign{\smallskip}
    \bottomrule
    \end{tabular}
\vspace{-5mm}
\end{table}
\setlength{\tabcolsep}{4pt}

\vspace{2pt}\noindent\textbf{Exposure Time and FPS.}
We further study the effects of synthesized blurry video parameters for video deblurring, \eg, exposure time and FPS. To achieve so, we further synthesize five new deblurring datasets with different exposure times and the FPS~(by changing $T$ and $\tau$). Tab.~\ref{tab:shooting_param} demonstrate that both exposure time and FPS play significant roles in video deblurring for the following reasons. The exposure time of the blurry video reflect the degree of the blurs to some extent. Meanwhile, frame-rate affects the temporal modeling of the deblurring models. As a result, we should synthesize the blurry video for model training with closer video parameters of real-world blurry videos to achieve better deblurring performance.

\section{Conclusion and Limitation}
\label{sec:conclusions}
In this paper, we explore the real-world video deblurring task by synthesizing realistic blurs in the RAW space with noise. A novel blur synthesis pipeline RAW-Blur and a corresponding ultra-high frame-rate RAW video dataset UHFRaw are proposed. With numerous experimental results, existing video deblurring models trained with the blur dataset synthesized by RAW-Blur can significantly improve real-world video deblurring performance compared to existing blur synthesis pipelines. Meanwhile, we highly recommend synthesizing realistic blurs in RAW space with high FPS RAW frames and considering the effects of ISP and blurry video parameter settings. However, our pipeline cannot be applied to the existing synthetic datasets since they do not provide the RAW sources. One feasible solution is to invert the RGB frames into RAW images and use our RAW-Blur synthesis pipeline.

\section{Acknowledgement}
This work was supported partially by the Major Research Plan of the National Natural Science Foundation of China (Grant No. 61991450), the Shenzhen Key Laboratory of Marine IntelliSense and Computation (under Contract ZDSYS20200811142605016), and JSPS KAKENHI (Grant Number 22H00529).

\clearpage
% ---- Bibliography ----
%
% BibTeX users should specify bibliography style 'splncs04'.
% References will then be sorted and formatted in the correct style.
%
\bibliographystyle{splncs04}
\bibliography{egbib}

\newpage
\section*{Appendix}
\appendix
In this supplementary document, we provide additional materials to supplement our main submission. Specifically, in Sec.~\ref{sec:appendix1}, we provide additional examples of higher frame-rate sharp videos in the synthetic blur analysis part. Sec.~\ref{sec:appendix2} shows the proof of the denoising process by averaging several consecutive frames. Sec.~\ref{sec:appendix3} gives detailed evaluation metrics of the models trained on synthetic and real-world blur datasets. Meanwhile, more visual results on the real-world blurry scenarios in BSD~\cite{zhong2020efficient} and RSCD~\cite{zhong2021towards} are provided, which demonstrate the proposed realistic blur synthesis pipeline can help existing video deblurring models generalize well on real-world blurs. 

\section{Additional Examples of Ultra-high Frame-rate Sharp Video}
\label{sec:appendix1}
\begin{figure*}[htbp]
    \centering
    \includegraphics[width=0.98\linewidth]{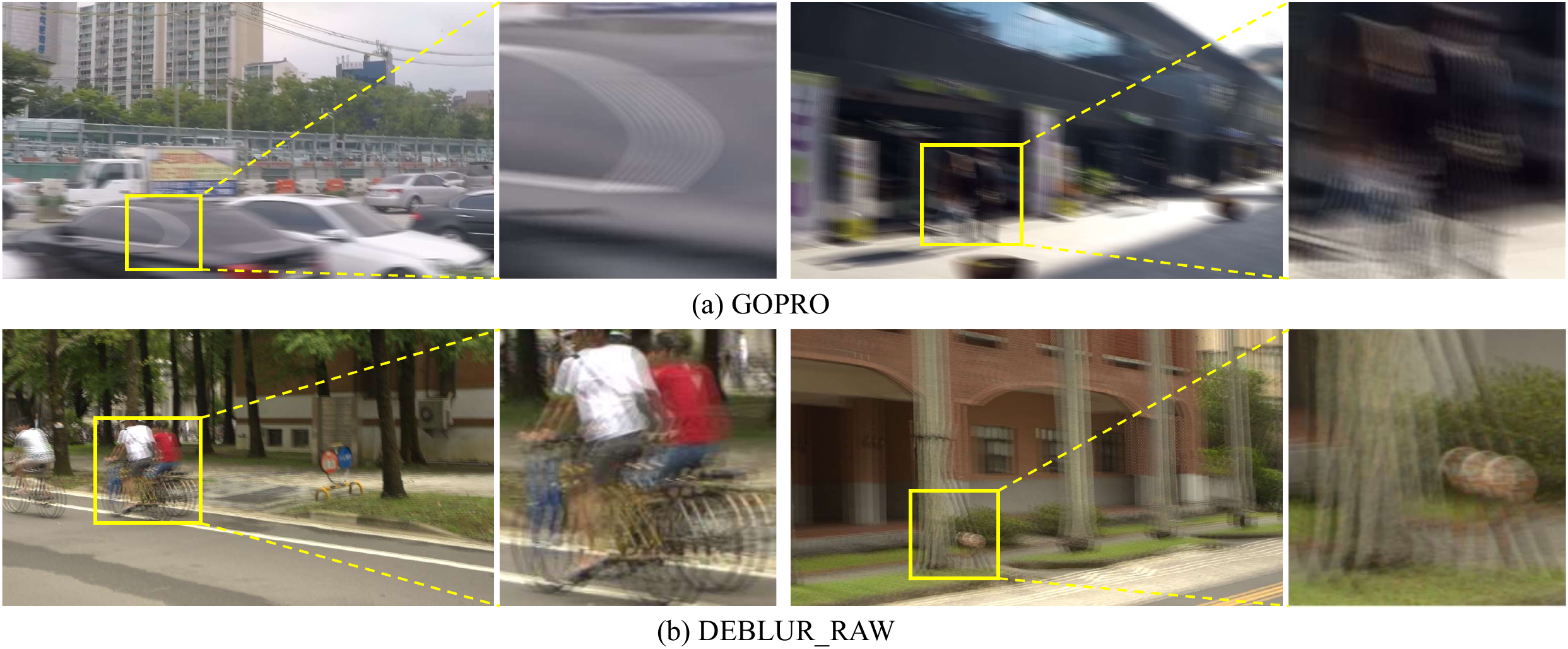}
    \vspace{-5mm}
    \caption{Artifacts of directly averaging low frame-rate sharp video frames in GOPRO~\cite{nah2017deep} and DEBLUR\_RAW~\cite{liang2020raw} deblurring datasets. }
    \label{fig:low_framerate_artifacts}
\end{figure*}

As we analysed in Sec. 3.3 of our main submission, we propose to capture higher frame-rate sharp videos to synthesize realistic blurs without undesired artifacts. Here, we further provide some synthesized blur examples in GOPRO~\cite{nah2017deep} and DEBLUR\_RAW~\cite{liang2020raw} that directly average sharp videos at 240fps and 30fps to synthesize blurs, respectively, resulting in discontinuous artifacts in the generated blurs shown in Fig.~\ref{fig:low_framerate_artifacts}. These artifacts greatly deteriorate the real-world video deblurring performance. 

\begin{figure*}[htbp]
    \centering
    \vspace{-5mm}
    \includegraphics[width=0.90\linewidth]{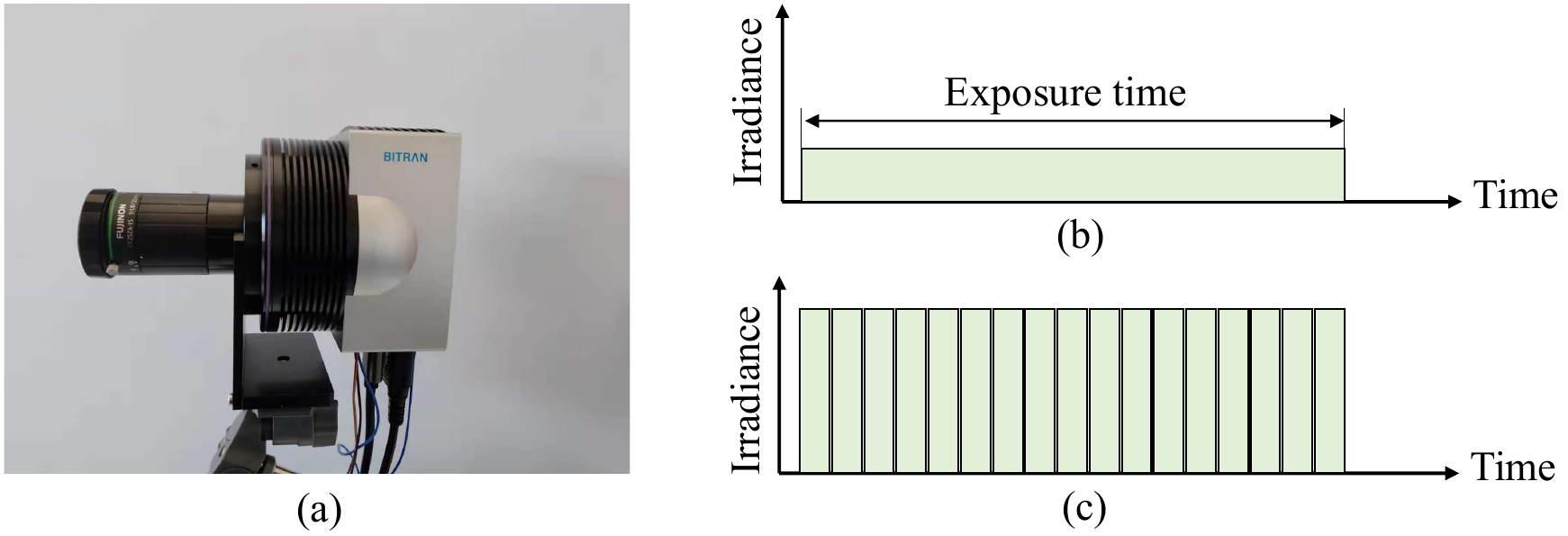}
    \vspace{-5mm}
    \caption{The acquisition device and exposure schedule. \textbf{(a)} BITRAN CS-700C Camera. \textbf{(b)} Real-world blur accumulation exposure schedule. \textbf{(c)} Our exposure schedule for capturing higher framerate sharp videos. Thanks to the very high frame-rate and little inter-frame time~(dead time), we can synthesize realistic blurs without aforementioned artifacts. }
    \label{fig:camera_exposure}
\end{figure*}

\begin{figure*}[htbp]
    \centering
    \vspace{-10mm}
    \includegraphics[width=0.98\linewidth]{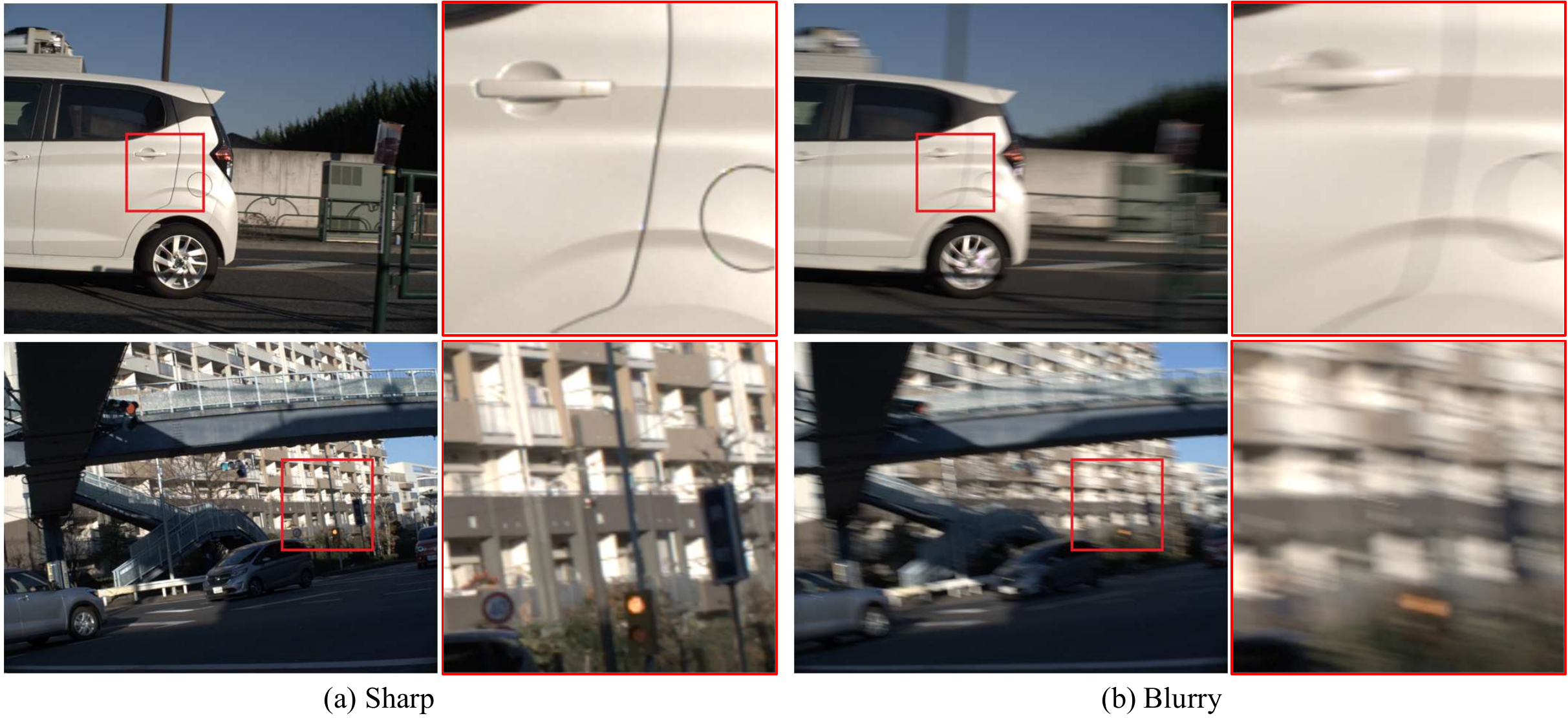}
    \vspace{-5mm}
    \caption{Sharp-blurry examples in the blur dataset synthesized with UHFRaw. We see that the blurs are realistic. }
    \label{fig:synthesized_blur_example}
\end{figure*}

Compared to these existing datasets, the sharp videos in UHFRaw dataset are captured at 940 fps, and the exposure time is 1.01 ms~(the camera and exposure schedule are shown in Fig.~\ref{fig:camera_exposure}). Thus the frames in UHFRaw are sharp and can serve as the ground truth for training. Meanwhile, we can synthesize continues blurs without aforementioned artifacts with these high frame-rate sharp frames. Some examples of the synthesized blur-sharp pairs with UHFRaw dataset are further provided in Fig.~\ref{fig:synthesized_blur_example}.


\section{Proof of Noise Reduction in Averaging Process}
\label{sec:appendix2}
As demonstrated in~\cite{foi2008practical,bielova2019digital}, the noise in camera imaging process actually consist of two parts, \ie~signal-dependent noise and signal-independent noise, which are respectively dependent on the input signal and the properties of sensors. The first part is usually the shot noise, which obeys the Poisson distribution and mainly comes from the photon-counting process to transfer the photons into electrons. The second part, which follows the Gaussian distribution, is the summation of several signal-independent noise, including read noise, row noise, \etc.

Thus, the obtained signal can be formulated as follows:
\begin{equation}
    \label{eq:noise_model}
    I = \hat{I} + N =  \hat{I} + N_{in} + N_{de},
\end{equation}
where $I$ is the obtained noisy signal; $\hat{I}$ is the unknown original signal; $N_{in}$ is the signal-independent noise part, and $N_{de}$ is the signal-dependent noise part. The expectations are zero~\cite{foi2008practical}, and the variance of them can be combined into a affine form~\cite{foi2008practical}:
\begin{equation}
    \label{eq:affine}
    Var(N) = a\hat{I} + b,
\end{equation}
where $N$ is the total noise, and $a$, $b$ are the coefficients for signal-dependent and independent noise parts. 

We further denote $\{I_1, I_2, \cdots, I_n\}$ as $n$ consecutive sharp frames following the Eq.~\ref{eq:noise_model}. Thus when we average these frames to synthesize blurry frame $B$:
\begin{equation}
    B = \frac{1}{n}\sum_{i=1}^{n} I_i = \frac{1}{n}\sum_{i=1}^{n} \hat{I}_i + \frac{1}{n}\sum_{i=1}^{n}N_i = \hat{B} + N_B,
\end{equation}
where $\hat{B}$ is the ideal blurry signal without noise, and $N_B$ is the reduced noise part. Then the expectation and variance the noise can be formulated as:
\begin{equation}
    \label{eq:variance}
    E(N_B) = 0, \\
    Var(N_B) = Var(\frac{1}{n}\sum_{i=1}^{n}N_i) = \frac{1}{n^2}\sum_{i=1}^{n}a\hat{I}_i + b=\frac{a\hat{B}+b}{n},
\end{equation}

From Eq.~\ref{eq:variance}, the variance of the noise in the averaged blurry frame decreases to about $\frac{1}{n}$ of original captured sharp frame, resulting in that the synthesized blurs are with little noise compared to real blurry scenarios. In conclusion, averaging multiple frames can further reduce the noise. 

\begin{figure}[htbp]
    \centering
    \includegraphics[width=\linewidth]{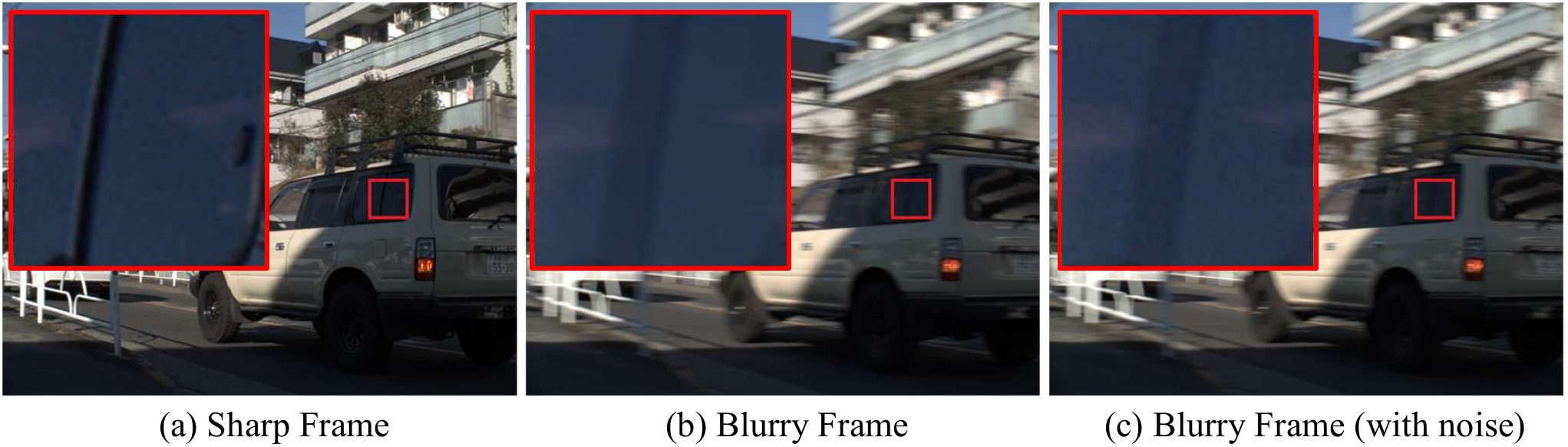}
    \caption{Visualization of the noise in different sharp and blurry frames. \textbf{(a)} One of the sharp frame used to synthesize blurry frame. \textbf{(b)} Synthesized blurry frame from (a). \textbf{(c)} Synthesized blurry frame from (a) with Gaussian noise. We see that directly averaging consecutive frame will reduce the real-world noise. }
    \label{fig:noise_visulization}
\end{figure}

Consequently, as illustrated in Eq.~4 in the main submission, we further add the reduced noise into the synthesized blurry frame to simulate real-world blurry scenarios. We further provide some visual comparison of the noise in the sharp frame and synthesized blurry frames in Fig.~\ref{fig:noise_visulization}. We see that the sharp frame also contains some noise due to the signal conversion process. Meanwhile, directly averaging frames would eliminate the noise shown in Fig.~\ref{fig:noise_visulization}(b). Accordingly, the noise is very close to the sharp frame when we add Gaussian noise into the synthesized RAW blurry frame shown in Fig.~\ref{fig:noise_visulization}(c). 

\section{Additional Quantitative and Qualitative Results}
\label{sec:appendix3}
\vspace{2pt}\noindent\textbf{More Quantitative Comparison Results. }
We further provide the detailed evaluation metrics~(\ie, PSNR and SSIM) of the models trained on different synthetic and real-world deblurring datasets, corresponding to Fig.~4 in our main submission. The results are shown in Tab.~\ref{tab:quantity_comparison}. We see that the model's performances surpass existing synthetic datasets with huge margin and are much closer to when trained on the training subset of BSD. These results demonstrate the proposed synthesis pipeline can generate more realistic and close-to-real blurs compared to existing pipelines. 

% ------- quantitative comparison ---------------
\setlength{\tabcolsep}{4pt}
\begin{table}[htbp] \footnotesize
    \centering
    \caption{Quantitative comparisons against existing synthetic blur datasets and the datasets synthesized by the proposed pipeline. }
    \label{tab:quantity_comparison}
    \resizebox{0.99\linewidth}{!}{
    \begin{tabular}{lcccccc}
    \toprule
    \noalign{\smallskip}
    Methods & GOPRO & DVD & REDS & HFR-DVD & Ours & BSD \\
    \noalign{\smallskip}
    \hline
    \noalign{\smallskip}
    DBN  & 24.47/0.7876 & 25.41/0.8114 & 25.41/0.8068 & 24.18/0.7982 & {31.28/0.9156} & 31.46/0.9166 \\
    EDVR & 24.32/0.7815 & 24.31/0.7816 & 24.45/0.7994 & 22.99/0.7820 & {31.68/0.9192} & 32.06/0.9158 \\
    \bottomrule
    \end{tabular}
    }
\end{table}

\begin{figure}
    \centering
    \begin{tabular}{c}
    \includegraphics[width=\linewidth]{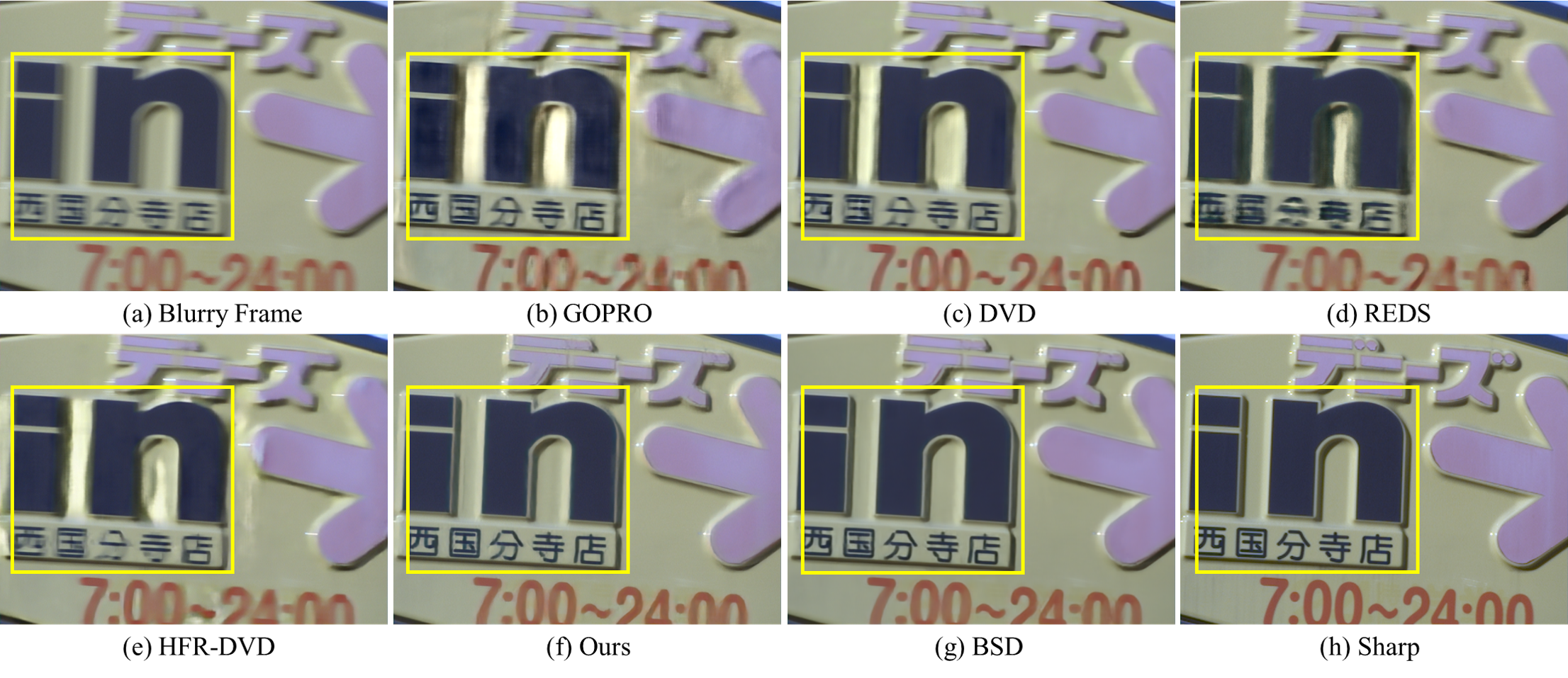} \\
    \includegraphics[width=\linewidth]{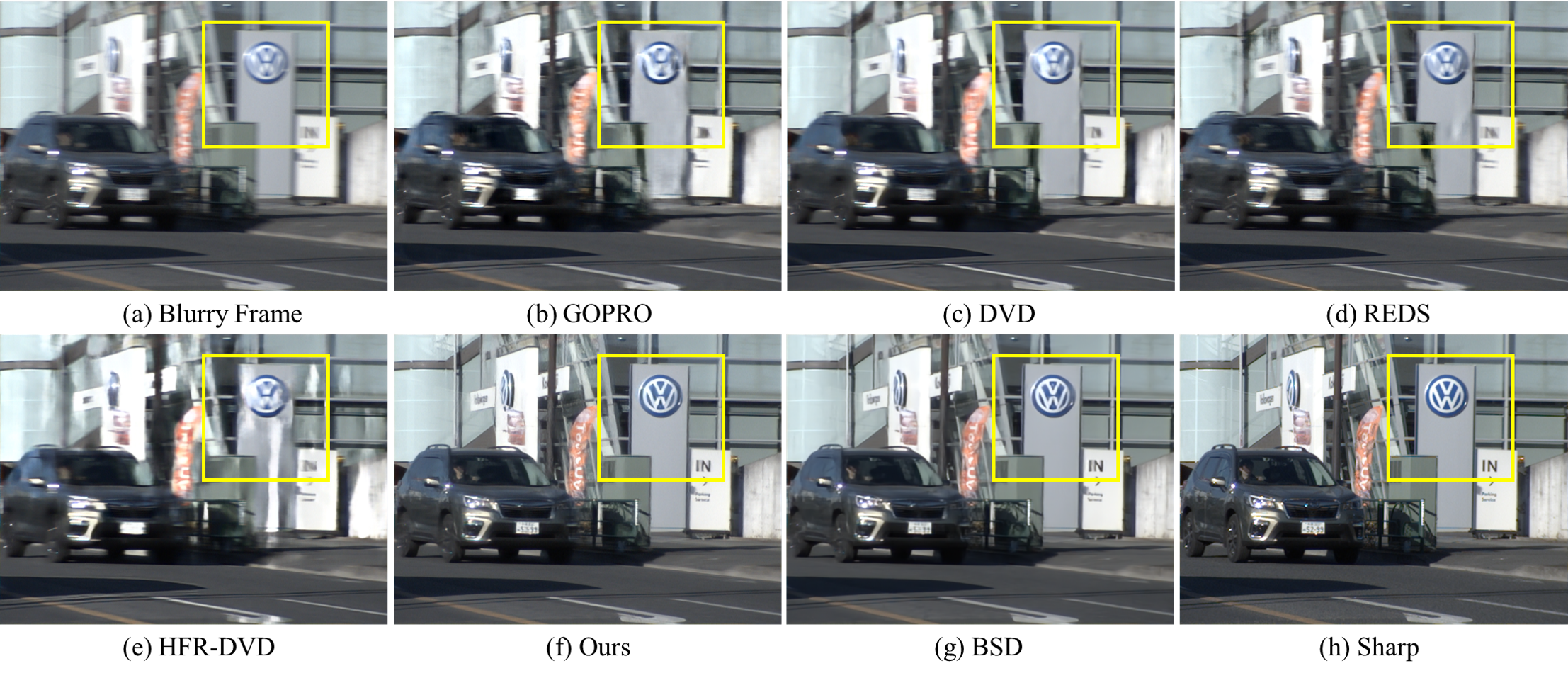} \\
    \includegraphics[width=\linewidth]{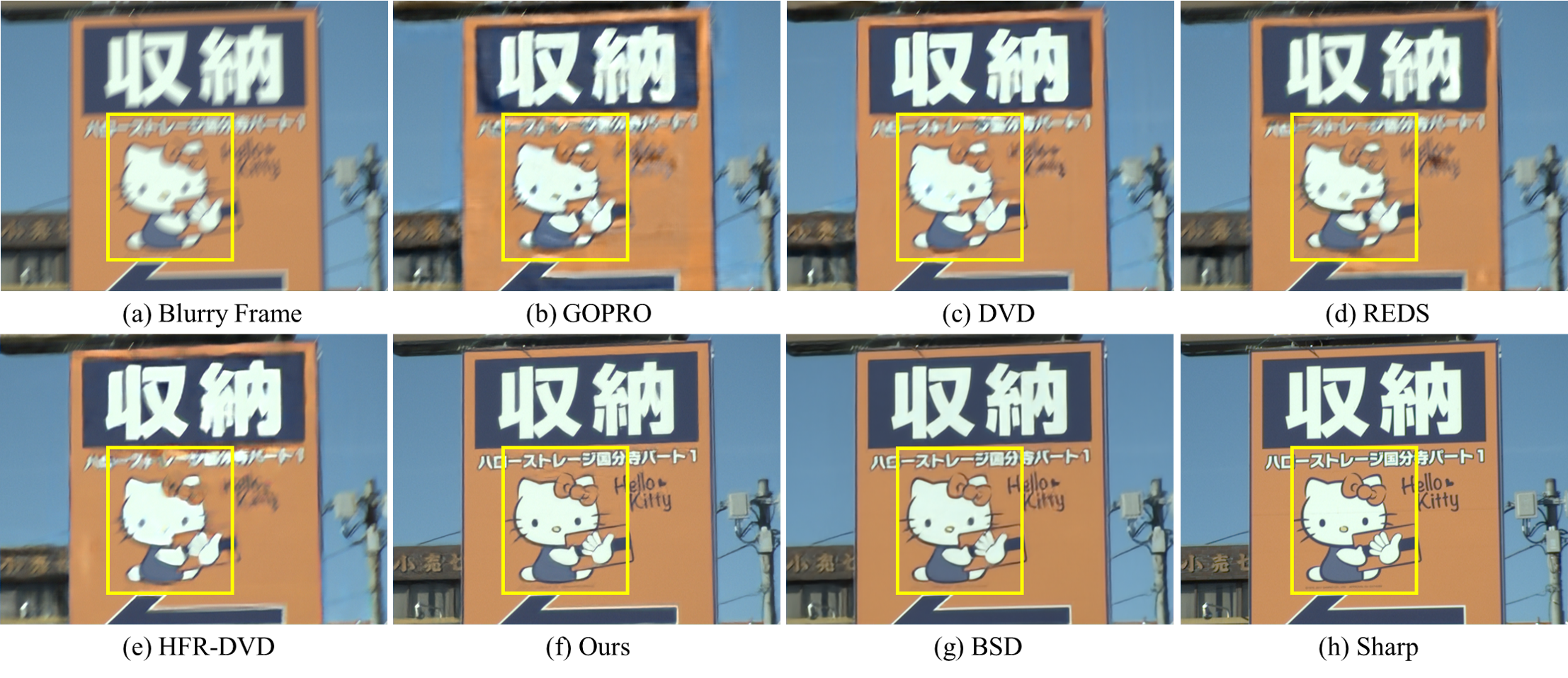} \\
    \end{tabular}
    \caption{More visual comparisons of DBN~\cite{su2017deep} on the testing subset of BSD (2ms-16ms)~\cite{zhong2020efficient}. The model trained on the dataset synthesized by our RAW-Blur pipeline achieved visual friendly results against that trained on existing synthetic video deblurring datasets. }
    \label{fig:visual_comparison_dbn}
\end{figure}

\begin{figure}
    \centering
    \begin{tabular}{c}
    \includegraphics[width=\linewidth]{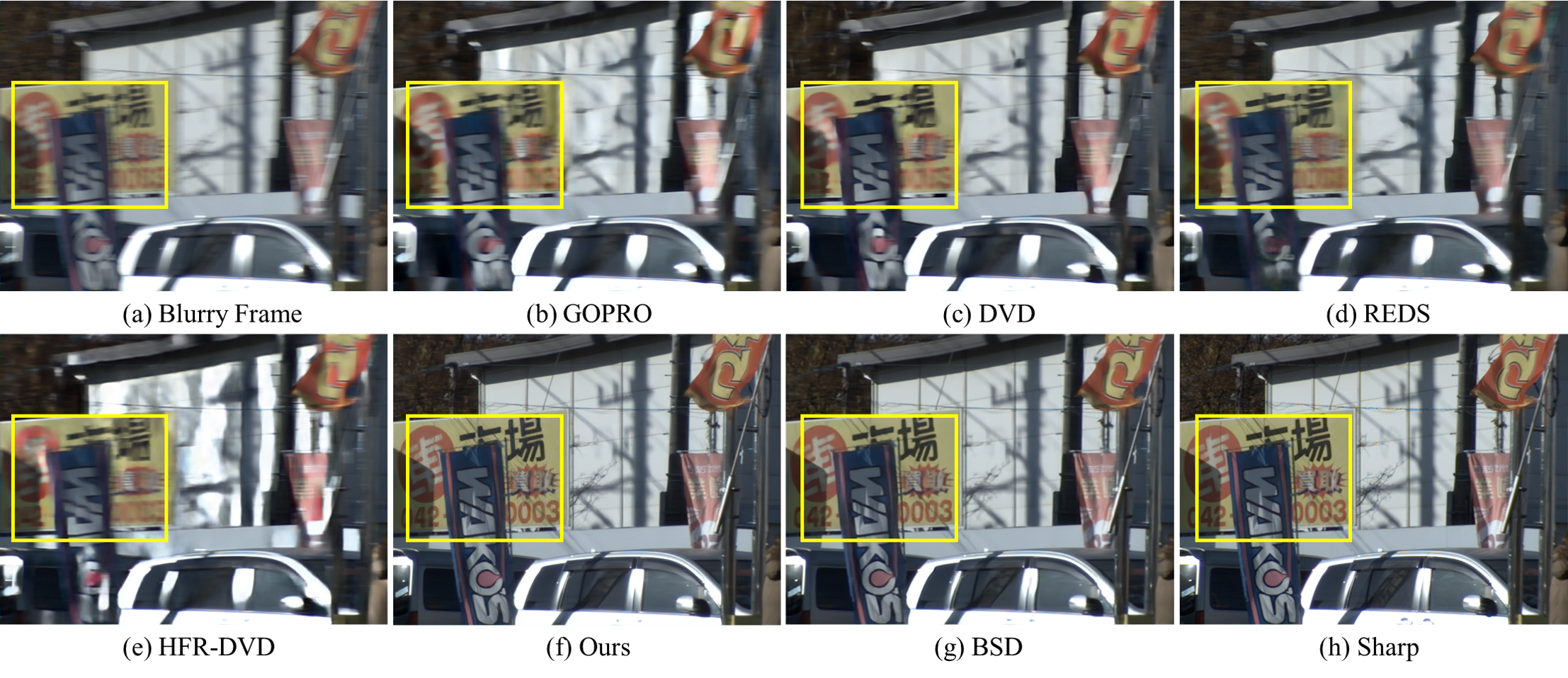} \\
    \includegraphics[width=\linewidth]{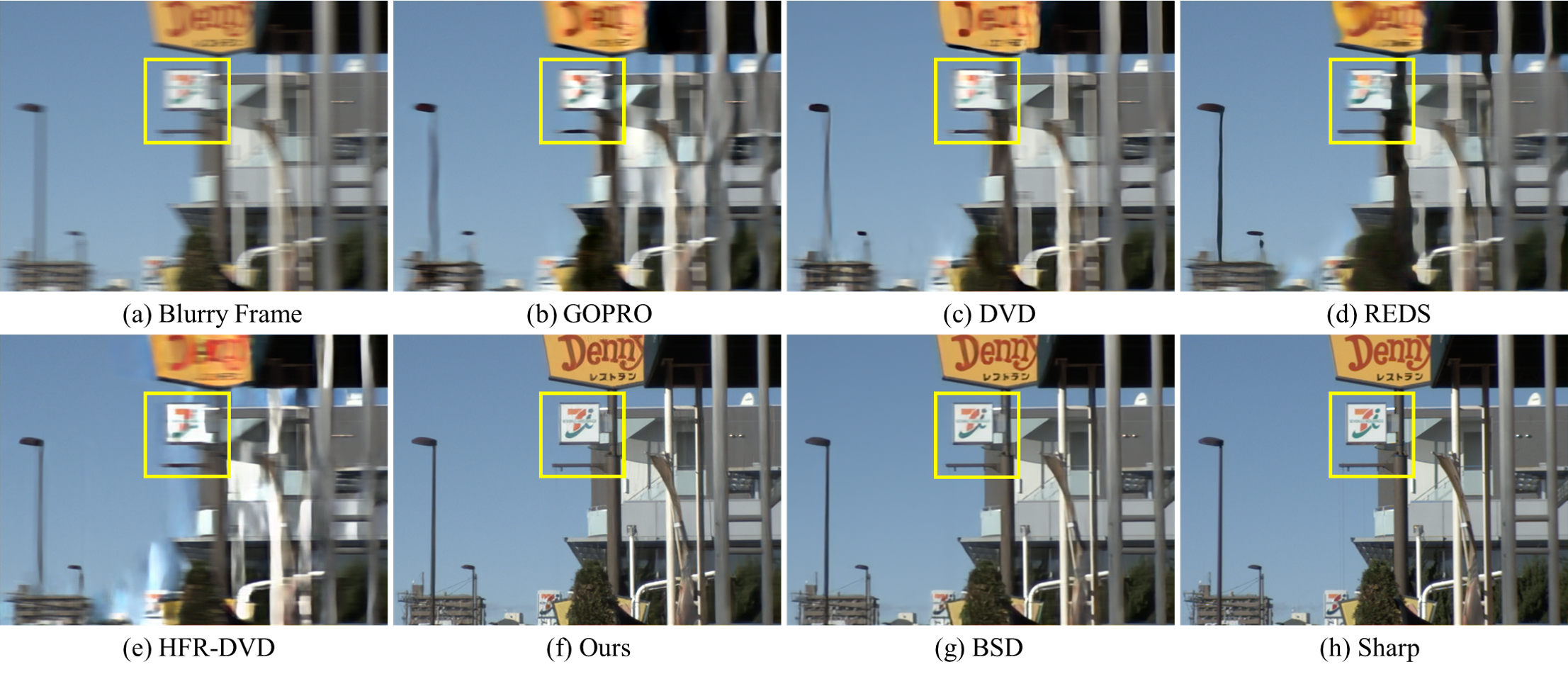} \\
    \includegraphics[width=\linewidth]{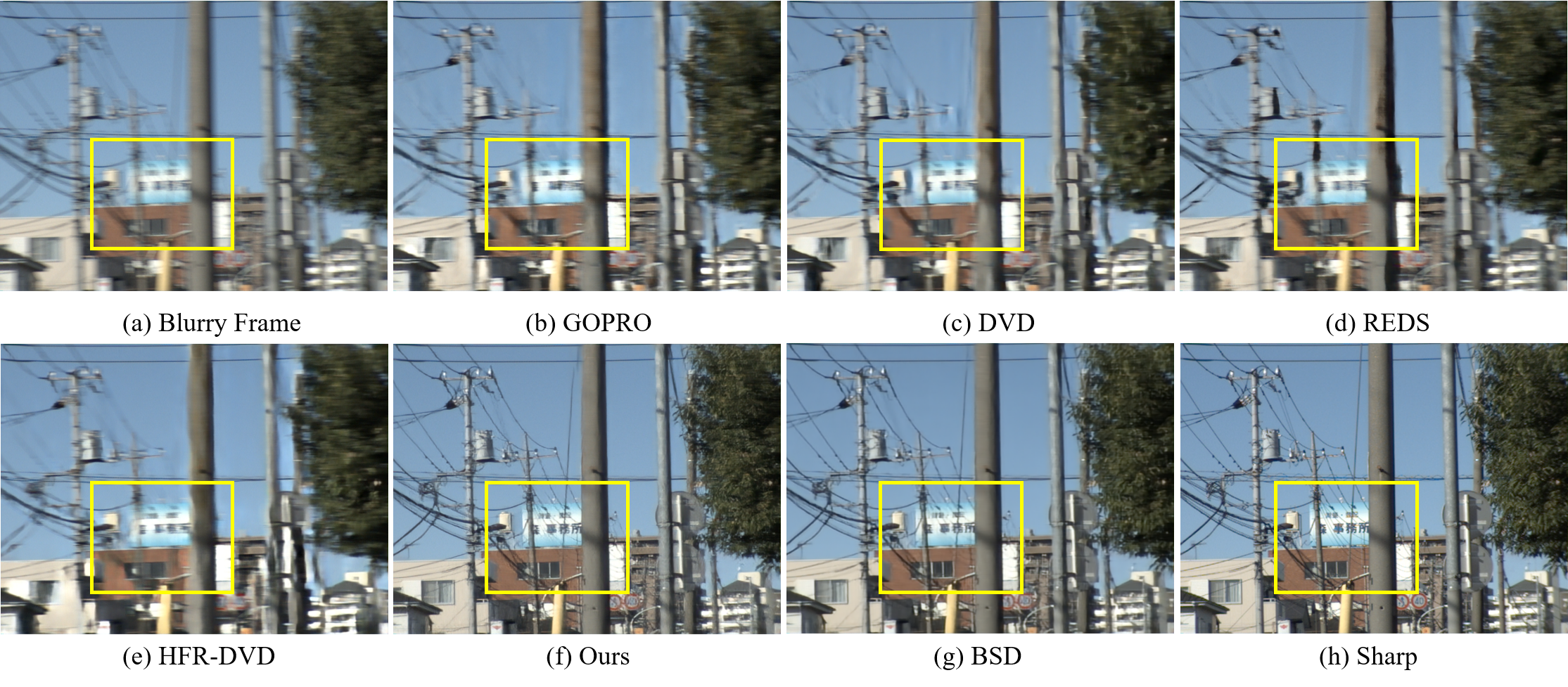}
    \end{tabular}
    \caption{More visual comparisons of EDVR~\cite{wang2019edvr} on the testing subset of BSD (2ms-16ms)~\cite{zhong2020efficient}. The model trained on the dataset synthesized by our RAW-Blur pipeline achieved visual friendly results against that trained on existing synthetic video deblurring datasets. }
    \label{fig:visual_comparison_edvr}
\end{figure}

\vspace{2pt}\noindent\textbf{More Qualitative Results on Real-World Blurry Scenes in BSD~\cite{zhong2020efficient}. }
We further provide some qualitative deblurred results of DBN~\cite{su2017deep} model and EDVR~\cite{wang2019edvr} model in Fig.~\ref{fig:visual_comparison_dbn} and~\ref{fig:visual_comparison_edvr}. We see that the models trained on existing synthetic datasets~(\ie, GOPRO, DVD, REDS, HFR-DVD) cannot deal with the blurs in the real-world blurry scenarios since the synthetic blurs owns completely different blur patterns compared to real-world blurs. On the contrary, these models trained on the blur dataset synthesized with our pipeline achieve visual friendly deblurred results. These results further demonstrate the effectiveness of the proposed blur synthesis pipeline. 

\begin{figure}
    \centering
    \begin{tabular}{c}
    \includegraphics[width=\linewidth]{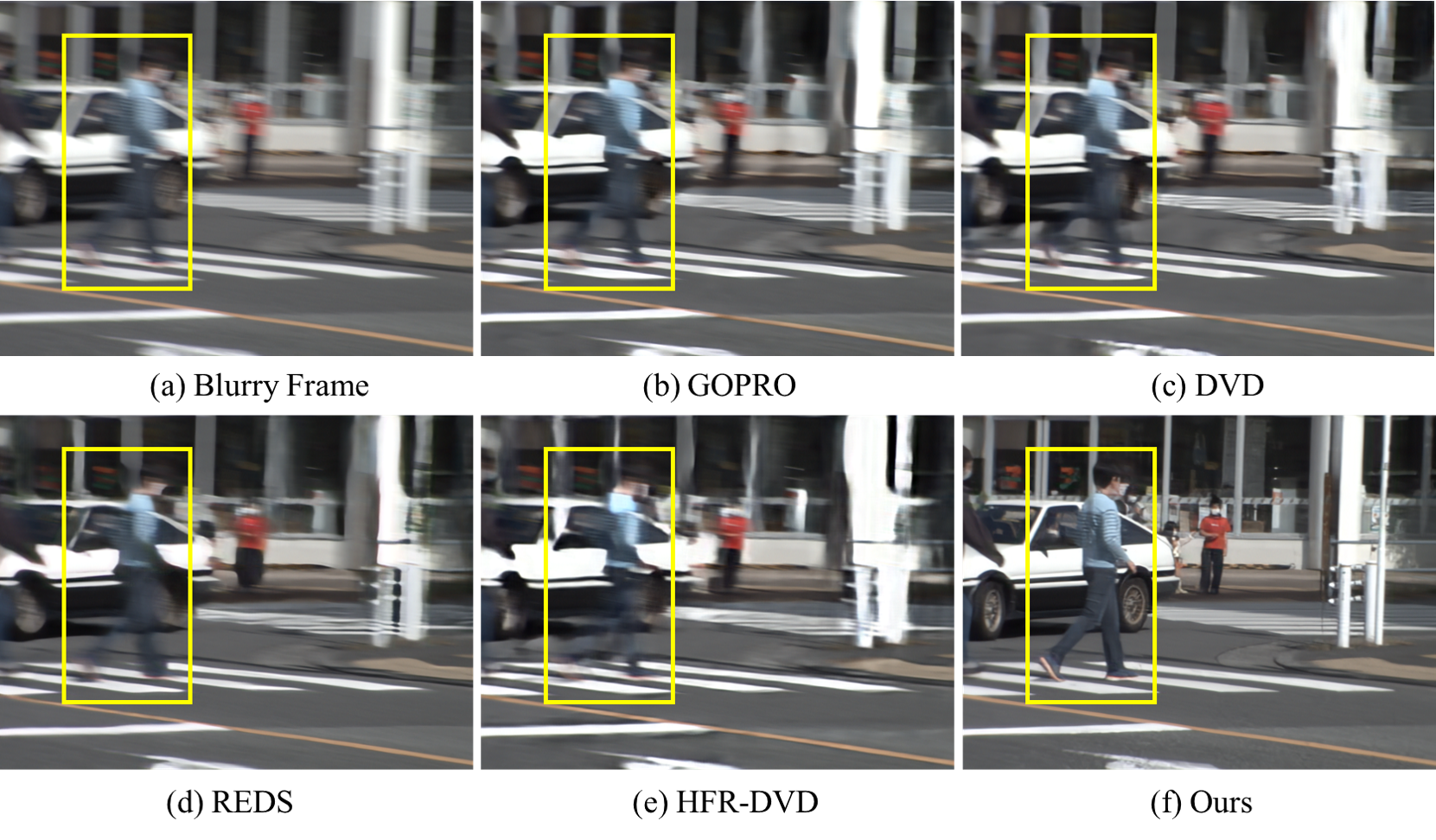} \\
    \vspace{5mm}
    \includegraphics[width=\linewidth]{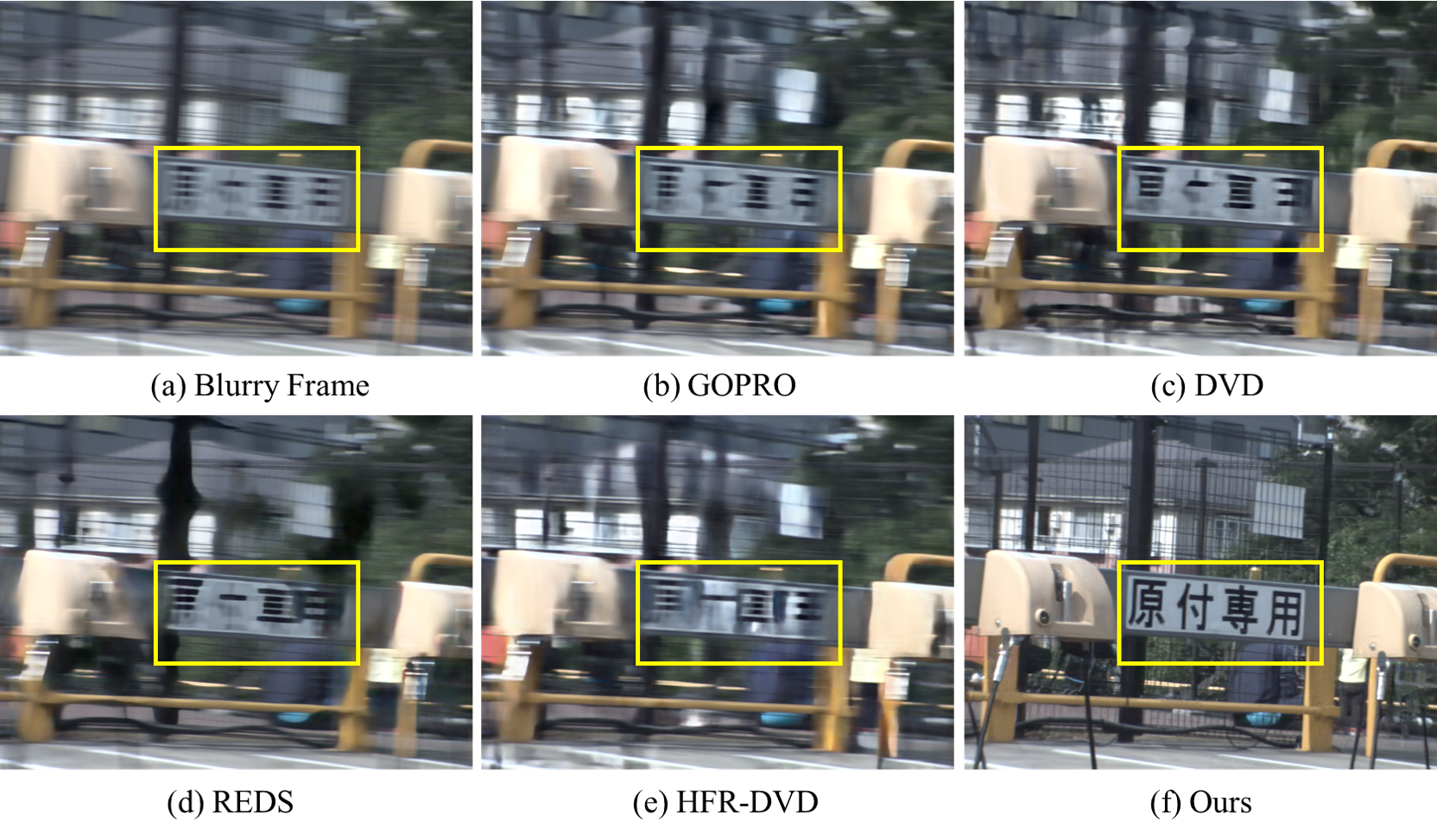} \\
    \end{tabular}
    \caption{More visual comparisons of EDVR~\cite{wang2019edvr} on the collected real-world blurry scenarios in RSCD~\cite{zhong2021towards}. }
    \label{fig:visual_comparison_rscd_edvr}
\end{figure}

\vspace{2pt}\noindent\textbf{More Qualitative Results on Real-World Blurry Scenes in RSCD~\cite{zhong2021towards}. }
We also test the proposed blur synthesis pipeline on the captured real-world blurry scenes in the Rolling Shutter Correction and Deblurring~(RSCD) dataset~\cite{zhong2021towards} which is captured by a rolling shutter camera with real-world blurs. The deblurred results of DBN and EDVR are shown in Fig.~\ref{fig:visual_comparison_dbn} and Fig.~\ref{fig:visual_comparison_rscd_edvr}, respectively. We should note that the models trained on existing synthetic datasets cannot deal with the blurs in these real-world blurry scenarios, while the models trained with data synthesized by the proposed pipeline can remove these blurs effectively.

\begin{figure}
    \centering
    \begin{tabular}{c}
    \includegraphics[width=\linewidth]{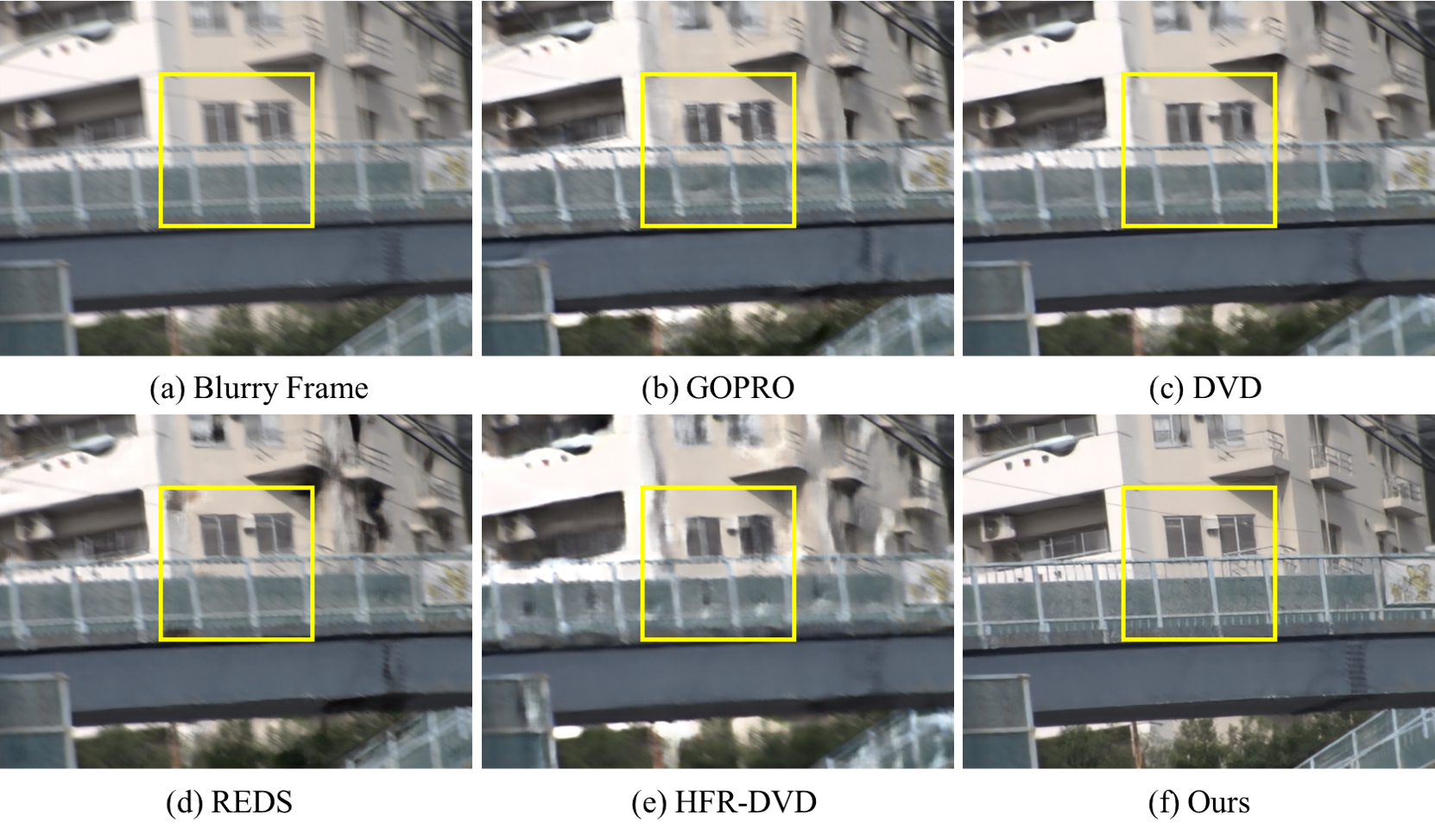} \\
    \vspace{5mm}
    \includegraphics[width=\linewidth]{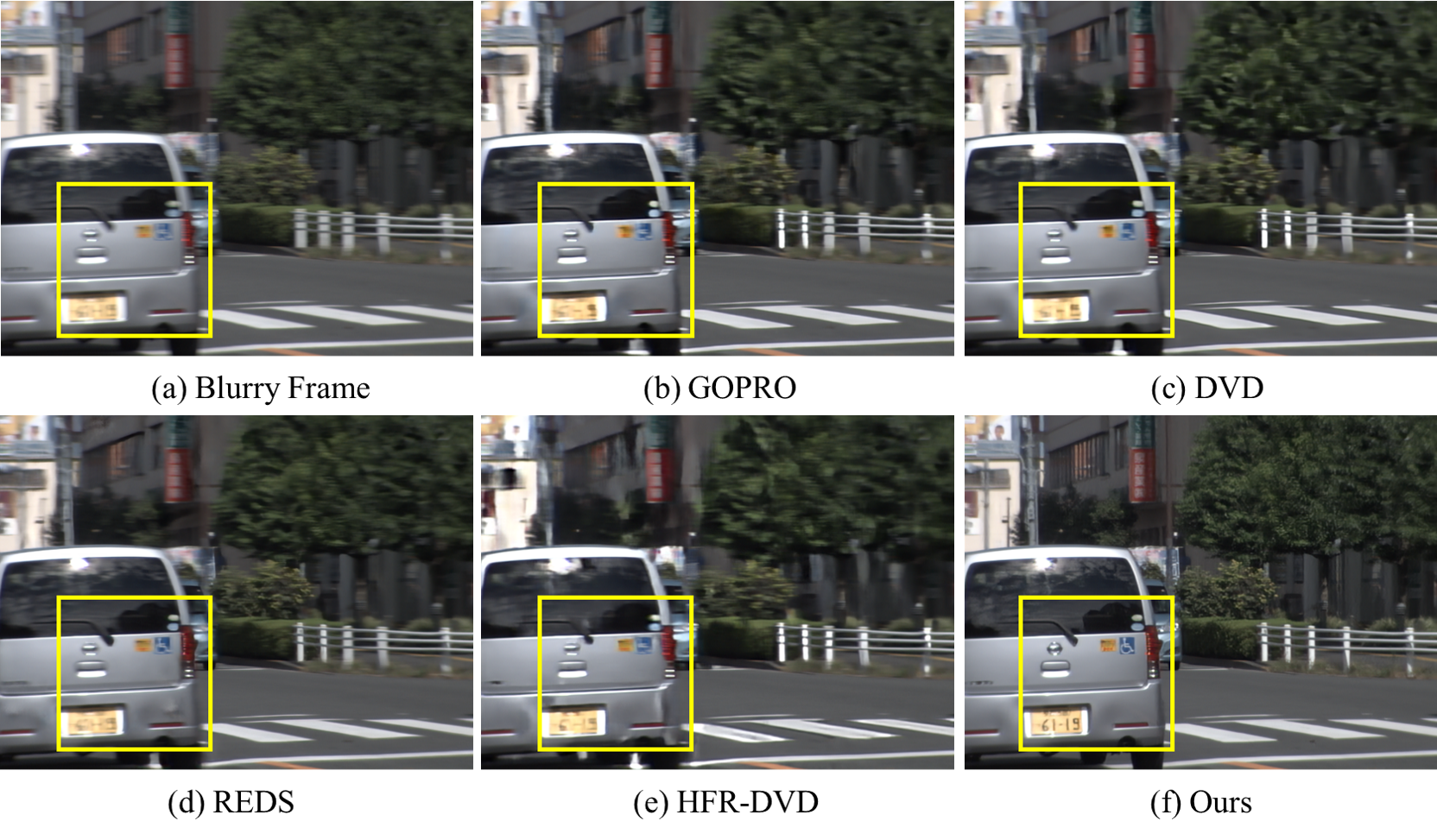} \\
    \end{tabular}
    \caption{More visual comparisons of DBN~\cite{su2017deep} on the collected real-world blurry scenarios in RSCD~\cite{zhong2021towards}. }
    \label{fig:visual_comparison_rscd_dbn}
\end{figure}

\end{document}